\documentclass[11pt, a4paper, logo]{mll_style}

\usepackage{natbib}
\usepackage{adjustbox}
\usepackage{array}
\usepackage{float}
\usepackage{placeins}
\usepackage{enumitem}
\usepackage{multirow}
\usepackage{makecell}
\usepackage{mathtools}
\usepackage{amsmath}
\usepackage{amssymb}
\usepackage{amsfonts}
\usepackage{subcaption}
\usepackage{booktabs}
\usepackage{tabularx}
\usepackage{rotating}
\usepackage{nicefrac}
\usepackage{pifont}
\usepackage{etoolbox}
\usepackage{needspace}
\usepackage[most]{tcolorbox}
\usepackage{caption}
\usepackage[capitalize,noabbrev]{cleveref}
\usepackage{titletoc}
\usepackage[toc,page,header]{appendix}

\definecolor{MyBlue}{rgb}{0.46, 0.50, 0.61}

\newcommand{\compactequationstyle}{%
  \setlength{\abovedisplayskip}{3pt}%
  \setlength{\belowdisplayskip}{3pt}%
  \setlength{\abovedisplayshortskip}{2pt}%
  \setlength{\belowdisplayshortskip}{2pt}%
}
\AtBeginEnvironment{equation}{\compactequationstyle}
\AtBeginEnvironment{equation*}{\compactequationstyle}
\AtBeginEnvironment{align}{\compactequationstyle}
\AtBeginEnvironment{align*}{\compactequationstyle}

\newcommand{\cmark}{\textcolor{green!60!black}{\ding{51}}}
\newcommand{\xmark}{\textcolor{red!70!black}{\ding{55}}}
\newcommand{\pmark}{\textcolor{orange!80!black}{$\sim$}}

\graphicspath{{figures/}}

\reportnumber{001}
\paperurl{https://ragen-ai.github.io/bagen}

\title{\centering BAGEN: Are LLM Agents Budget-Aware?\vspace{-8pt}}

\date{}

\author{{\normalsize
Yuxiang Lin$^{12*}$,
Zihan Wang$^{12*\dagger}$,
Mengyang Liu$^{3*}$,
Yuxuan Shan$^{2*}$,
Longju Bai$^{4*}$,
Junyao Zhang$^{2}$,
Xing Jin$^{3}$,
Boshan Chen$^{2}$,
Jinyan Su$^{5}$,
Xingyao Wang$^{6}$,
Jiaxin Pei$^{78}$,
Manling Li$^{1}$\\
{\small $^*$Core contributors. \quad $^\dagger$Project lead.}\\
{\small\mdseries $^1$Northwestern University \quad $^2$O2 Lab \quad $^3$Independent \quad
$^4$University of Michigan \quad $^5$Cornell \quad
$^6$All~Hands~AI \quad $^7$Stanford \quad $^8$UT Austin}\\
{\small \url{https://ragen-ai.github.io/bagen}}
\vspace{-10pt}
}}

\begin{abstract}
While agents are increasingly spending more resources, today agent cost is mostly measured only after execution. A \textbf{Budget-Aware Agent (BAGEN)} should treat budget as an active control signal, rather than a passive cost metric. We first systematically define budget estimation as internal budgets (from agent computation) and external budgets (from agent actions). We then formalize budget-awareness as \textbf{progressive interval estimation}: at each step of a plan, an agent should predict an upper and lower bound on remaining budget, and alert when completion is unlikely.
Scoring with a rollout-replay protocol, we find consistent failure patterns on four environments and five frontier agents: (1) strong agents do not necessarily have strong budget-awareness, with correlation $r \approx 0.35$. (2) frontier models are consistently over-optimistic, continue spending on tasks that are unlikely to succeed, instead of alerting the user early. (3) budget-aware signal is actionable and trainable. Early stop saves 28--64\% tokens on failed trajectories, and SFT+RL strengthens early stop and alert behavior. (4) precise interval calibration remains challenging, with interval coverage capping at 47\% after SFT+RL.
\end{abstract}

\begin{document}

\maketitle

%
%

 \begin{abstract}
  Foundation-model agents are deployed with growing resource constraints like token, money, and time budgets, yet it remains unclear whether they know how much budget they will spend. We call this capability
  \emph{budget awareness} and formalize an ability for Budget-Aware Agents (BAGEN) as \emph{progressive interval estimation}:
  mid-execution, whether the agent can provide an interval on how much budget remain needed and whether the task is still finishable.
  We score this with a rollout-replay protocol that re-queries the same agent on every prefix of an unconstrained rollout, and decompose estimation into three sub-capabilities: feasibility prediction, early failure detection, and interval
  calibration. We evaluate five frontier models on four environments spanning \emph{internal} budgets (token consumption on Sokoban, Search-R1, and SWE-bench) and \emph{external} budgets (cost, time, and warehouse occupancy in a supply-chain environment curated from real enterprise data); we further train Qwen-7B budget estimators with SFT and RL on Sokoban, and deploy their predictions through a simple early-stop policy. Across these axes, we find budget awareness: (1) 
    decouples from task performance, (2) fails in structured ways, and (3) is already actionable and trainable as a control signal that resource-limited agents currently lack. Code for this project will be open-sourced.
  \end{abstract}

\vspace{-8pt}
\section{Introduction}
\label{sec:v2:intro}

Foundation-model agents are increasingly deployed in longer horizons and higher-stakes tasks: a coding agent consumes tokens per reasoning step, a web agent spends API calls per search query, and a supply-chain agent commits real dollars and warehouse capacity per procurement decision. The budget their generation consumes (their \emph{internal budget}, primarily tokens) and the budget their actions commit (their \emph{external budget}, including money, time, and inventory) are both growing rapidly with deployment horizon. Yet existing benchmarks track this budget only after the fact, rarely asking whether the agent itself knew, mid-execution, what it was about to spend.

\begin{figure}[t]
    \centering
    \includegraphics[width=1\linewidth]{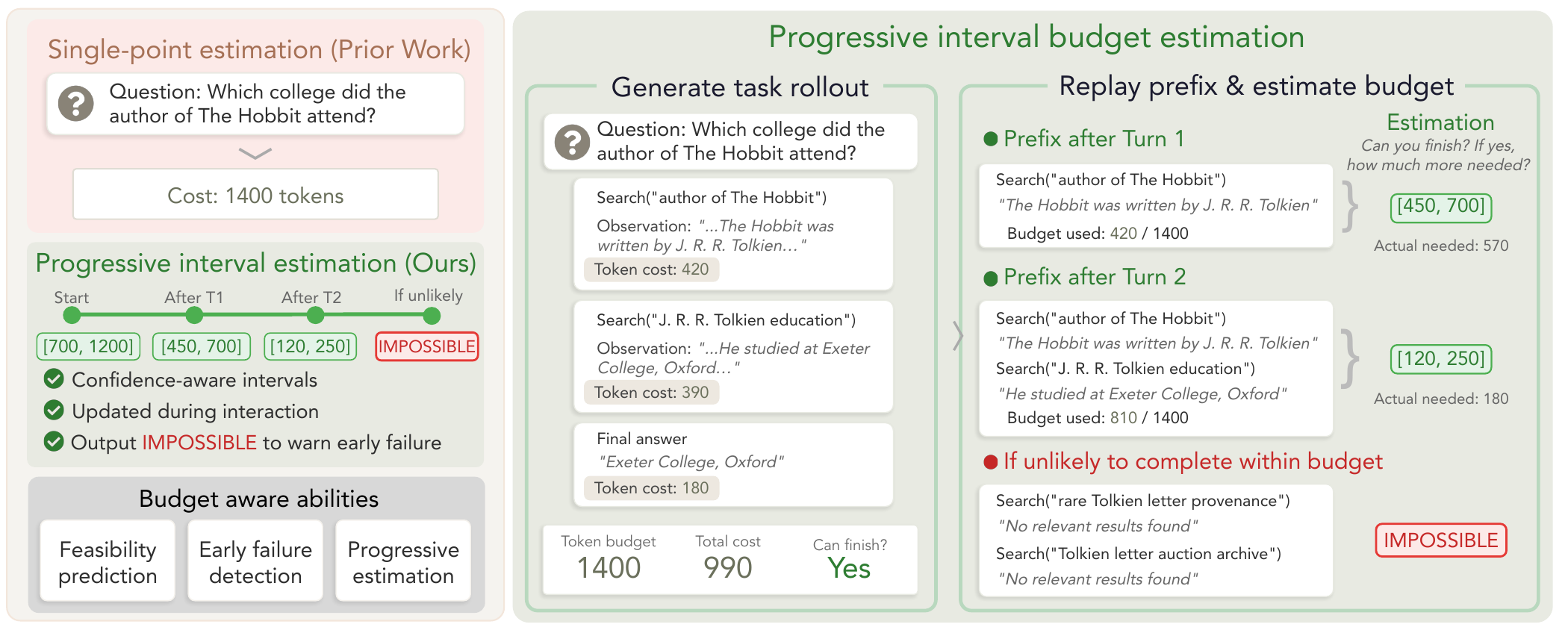}
    \caption{Progressive interval estimation in BAGEN. We record an unconstrained rollout, then re-query the same agent on every prefix to predict either an interval over remaining budget or an \texttt{impossible} declaration; predictions are scored against the realized remaining budget and outcome.}
    \label{fig:v2:overview}
    \vspace{-10pt}
\end{figure}

This brings up a fundamental question: \emph{does the agent know how much budget it needs?} We call this capability \emph{budget awareness}: the ability of a Budget-Aware Agent (BAGEN) to estimate, mid-execution, how much budget remains and whether the task is still finishable. An agent that cannot estimate its own resource requirements cannot decide when to abort a hopeless task, when to request more resources, or how to allocate budget across sub-goals.

Two gaps prevent systematic study of this capability. \textbf{First}, agent research~\citep{liu2026budgetconstrained,ding2026calibratethenact,mccleary2026quantifying} usually calculates token consumption as a post-hoc metric, but hardly ever asks whether the agent could \emph{self-estimate} how much budget it would need. \textbf{Second}, most evaluation protocols collect a single-point prediction at task start, which mismatches long-horizon agentic tasks where feasibility evolves turn by turn: a project manager re-estimates the remaining timeline at every milestone, gives a range rather than a point, and flags when completion becomes infeasible.

To address these limitations, we propose \emph{progressive interval estimation} as a new agent capability that is \textit{confidence-aware} and \textit{progressive throughout execution}. We record a full agent rollout without any budget constraint, then query the same agent separately at every turn: \emph{given current progress, how much budget remains to finish? Provide an interval with confidence, or declare the task impossible.} We decompose this capability into three sub-capabilities (feasibility prediction, early failure detection, and interval calibration) and ask five frontier models to perform it across four environments (Sokoban, Search-R1, SWE-bench, and a Warehouse environment with three coupled budget dimensions curated from real enterprise data) covering both internal and external budget modalities. Across these experiments, we find:
\begin{itemize}[leftmargin=*, itemsep=2pt, topsep=2pt]
\item \textbf{Budget awareness is a distinct capability from task performance, separated by interval calibration rather than feasibility prediction.} Task success correlates only weakly with interval hit rate ($r{\approx}0.35$). On Search-R1, Opus achieves the highest task success rate ($75.8\%$) but Sonnet produces better intervals ($36.5\%$ vs.\ $23.1\%$); no model dominates all three sub-capabilities (\S\ref{sec:v2:decoupling}).

\item \textbf{Binary feasibility is a calibration problem; interval estimation is a reasoning problem, and all models are optimistically biased.} SFT alone raises Qwen-7B feasibility accuracy from $25.5\%$ to ${\approx}90\%$, indicating the capability was latent; interval coverage only reaches $47\%$ after SFT+RL. All twenty model–environment pairs underestimate remaining budget more often than they overestimate it, and weaker models are \emph{more} optimistic, not less (\S\ref{sec:v2:why-fail:h1}, \S\ref{sec:v2:why-fail:h2}).

\item \textbf{Estimates update over turns, but failure is recognized too late to act on.} Estimates shift as execution progresses—but not reliably toward the truth. On failed trajectories, models predict feasibility above $70\%$ even after $60\%$ of the budget is consumed; the alarm fires only in the final $20\%$ (\S\ref{sec:v2:why-fail:h3}, \S\ref{sec:v2:why-fail:h4}).

\item \textbf{The signal is actionable but training is fragile.} An early-stop policy keyed on \texttt{impossible} predictions saves between $28\%$ and $64\%$ of tokens on failed trajectories at a cost of only $1.6$ to $4.2$ percentage points in success rate. SFT followed by RL improves estimation on Sokoban; RL without an SFT warm-start collapses entirely (\S\ref{sec:v2:use}).
\end{itemize}

In general, our results show that budget is less a metric for after-the-fact accounting and more a control signal that resource-limited agents currently lack.

\section{Two Failures of Single-Point Budget Estimation}
\label{sec:v2:naive-probe}
\vspace{-4pt}

We start with the simplest form of budget estimation: at task start, ask the agent for a single-point estimate of the total tokens it will spend, and score it against the realized rollout budget. As a pilot probe, we elicit such estimates from five frontier models on two internal-budget environments, Sokoban and Search-R1, collecting both a first-turn estimate (at task start) and later-turn estimates (after replaying logged prefixes);   full setup is in App.~\ref{app:pilot-setup}. Two failures hold across all models, and they motivate the protocol of \S\ref{sec:v2:method}.                                                                                  

\textbf{The first failure is systematic optimism.} Across all five models on both tasks, first-turn predictions underestimate the realized budget more often than overestimating it (Figure~\ref{fig:v2:optimism-full});    the bias tracks model confidence rather than task difficulty, with weaker models on a task being \emph{more} optimistic, not less. 

\textbf{The second failure is that first-turn and later-turn estimates do not agree.} On Sokoban, Gemini's feasibility macro-$F_1$ improves by $+21.9$ points from first-turn to all-turn evaluation; on Search-R1, Qwen3-235B moves the opposite way by $9.3$ points (Figure~\ref{fig:v2:first-vs-all}). First-turn judgment is therefore neither a consistent over- nor underestimate of all-turn judgment; it is just a different judgment, and which side it falls on is model- and task-specific.

Together these failures argue against single-point estimation. A point estimate cannot express the optimism the model is exhibiting, but an interval can, and an explicit \texttt{impossible} option lets the model declare    infeasibility rather than silently underestimating it. A single query at $k{=}0$ misses the refinement that comes from observing partial progress, but a query that fires every turn captures it. \S\ref{sec:v2:method}
formalizes both changes as \emph{progressive interval estimation}.

\vspace{-6pt}

\section{Agent Budget Awareness}
\label{sec:v2:method}
\vspace{-5pt}

The previous section surfaced two structural failures of single-point budget estimation: systematic optimism and instability between first-turn and later-turn estimates. We now formalize the capability for Budget-Aware Agents (BAGEN) that addresses both. We first introduce two budget modalities, then describe the progressive interval estimation protocol with its rollout-replay procedure, and finally decompose estimation quality into three sub-capabilities with concrete metrics. The section closes with the experimental setup.

\vspace{-6pt}

\subsection{Budget Modalities: Internal and External}
\label{sec:v2:method:modalities}

\vspace{-6pt}


Consider an agent executing a multi-turn trajectory $\tau = \{(o_t, z_t, a_t)\}_{t=1}^{T}$, where $o_t$ is the observation (e.g., environment state or tool output) received at turn~$t$, $z_t$ is the agent's reasoning trace, and $a_t$ is the action. At each turn, the agent commits budget in two fundamentally different ways.

\textbf{Internal budget.}~~
Internal budget is the compute generated by the model's own reasoning. Let $c_t^{\text{in}}$ denote the fresh token count at turn~$t$. The cumulative internal cost up to turn~$k$ is $C_k^{\text{in}} = \sum_{t=1}^{k} c_t^{\text{in}},$
and the remaining internal cost from turn~$k{+}1$ onward is $R_k^{\text{in}} = C_T^{\text{in}} - C_k^{\text{in}}$. The agent operates under a total cap $B^{\text{in}}$, requiring $C_T^{\text{in}} \leq B^{\text{in}}$. A trajectory that exceeds this cap is truncated and counted as a failure. Internal-budget estimation tests whether the model can predict its own compute footprint: how many tokens will the remaining reasoning, planning, and action steps cost?

\textbf{External budget.}~~
External budget is the cost the agent commits in the environment. It is generally multi-dimensional: let $\mathbf{c}_t^{\text{ex}} \in \mathbb{R}^D$ denote the $D$-dimensional cost vector at turn~$t$, with cumulative usage and remaining cost $  \mathbf{C}_k^{\text{ex}} = \sum_{t=1}^{k} \mathbf{c}_t^{\text{ex}},$ where $\mathbf{R}_k^{\text{ex}} = \mathbf{C}_T^{\text{ex}} - \mathbf{C}_k^{\text{ex}},$
subject to per-dimension constraints $C_T^{\text{ex},(d)} \leq B^{(d)}$ for $d = 1, \ldots, D$. External budget forces the agent to reason about coupled resource constraints: holding more inventory raises revenue but eats warehouse capacity, while drawing credit improves short-term cash but creates repayment pressure later.

\vspace{-6pt}

\subsection{Progressive Interval Estimation and Rollout-Replay}

\vspace{-6pt}

\label{sec:v2:method:protocol}

\textbf{Progressive interval estimation.}~~
At every turn~$k$, given the trajectory prefix $\tau_{\leq k} = \{(o_t, z_t, a_t)\}_{t=1}^{k}$ and cumulative usage $C_k$, the estimator returns
\begin{equation}
  \hat{y}_k =
  \begin{cases}
    [\hat{R}_{k}^{\text{lo}},\; \hat{R}_{k}^{\text{hi}}]
      & \text{if the agent predicts the task is still feasible}, \\[4pt]
    \texttt{impossible}
      & \text{if the agent predicts completion is no longer achievable}.
  \end{cases}
  \label{eq:v2:progressive-estimation}
\end{equation}
The output captures three properties simultaneously: \emph{uncertainty} (interval width); \emph{progressiveness} (the estimate updates each turn); and \emph{infeasibility awareness} (an explicit \texttt{impossible} option, enabling early stopping or rerouting). The first two address the two single-point failures; the third makes the signal actionable downstream.

\textbf{Rollout-replay protocol.}~~
To separate budget estimation ability from task completion ability, we use a two-phase procedure (Figure~\ref{fig:v2:overview}).
\textbf{(1)~Rollout generation.} The agent executes the task without any budget constraint. We log the full trajectory $\tau$ together with per-turn cost $c_t$ and final outcome.
\textbf{(2)~Prefix replay and estimation.} For each non-terminal turn $k \in \{1, \ldots, T{-}1\}$, we replay the logged prefix $\tau_{\leq k}$ as history, append a cumulative-usage summary, and ask the agent to estimate $\hat{y}_k$ via Eq.~\ref{eq:v2:progressive-estimation}. Each prediction is scored against true remaining cost $R_k = C_T - C_k$ and the true outcome.

A natural alternative is to let the agent estimate budget \emph{during} its own rollout, interleaving estimation with execution. We avoid this because estimation itself consumes tokens, which would conflate task-completion cost with self-assessment cost. Online estimation is left to future work.

\vspace{-6pt}
\subsection{Sub-Capability Metrics}
\label{sec:v2:method:metrics}
\vspace{-6pt}

\begin{figure}[t]
    \centering
    \includegraphics[width=1\linewidth]{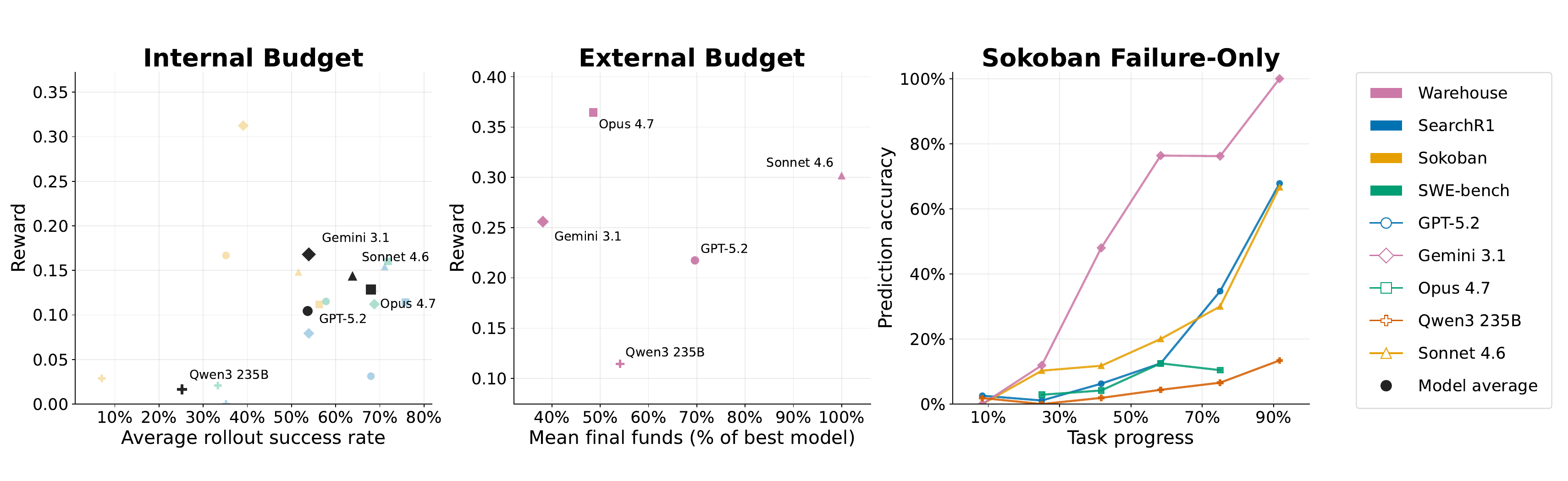}
    \caption{Left and middle: estimation quality is only weakly related to task performance for both internal and external budgets.
    Right: on failed Sokoban trajectories, estimation accuracy increases as more task progress is observed, with the largest gains appearing later in the trajectory.
    }
    \label{fig:v2:reward-vs-rollout-and-progress}
    \vspace{-10pt}
\end{figure}

A Budget-Aware Agent (BAGEN) must do three things: tell whether the task can succeed under budget, recognize failure early enough to act on it, and provide a calibrated cost range when success is possible. We score each as a separate sub-capability.

\textbf{(1) Feasibility prediction.}~~
\emph{Can the agent tell whether the task will succeed under the remaining budget?}
This is the coarsest, binary level of budget awareness. Let $y_k \in \{\text{feasible}, \text{impossible}\}$ be the ground-truth label at turn~$k$ (whether the agent succeeds within the budget), and $\hat{y}_k \in \{[\cdot,\cdot],\; \texttt{impossible}\}$ the model's output. We treat any interval as a ``feasible'' prediction and report
\begin{equation}
{
  \text{Macro-}F_1 = \tfrac{1}{2}\big(F_1^{\text{feasible}} + F_1^{\text{impossible}}\big),
  \label{eq:v2:macro-f1}}
\end{equation}
computed over the full trajectory (\emph{all-turn}) or the first turn only (\emph{first-turn}).

\textbf{(2) Early failure detection.}~~
\emph{For tasks that ultimately fail, can the agent recognize that early?}
This is critical for budget control: catching failure early makes early stopping and budget saving possible. Restricted to the positive class $y_k = \texttt{impossible}$,
\begin{equation}
{
  \text{Fail-}F_1 = \frac{2 \cdot \text{Prec}_{\text{imp}} \cdot \text{Rec}_{\text{imp}}}{\text{Prec}_{\text{imp}} + \text{Rec}_{\text{imp}}}.
  \label{eq:v2:fail-f1}}
\end{equation}
A high Fail-$F_1$ means the alarm fires when failure is real and stays silent on feasible trajectories.

\textbf{(3) Remaining budget estimation.}~~
\emph{When the task actually succeeds, can the agent predict that success and provide an accurate cost range?}
Restricted to trajectories that ultimately succeed, we score interval quality only when the model also correctly predicts feasibility. An \texttt{impossible} prediction on a successful trajectory scores zero:
\begin{equation}
{
  S_k =
  \begin{cases}
    \underbrace{\mathbf{1}\!\big[R_k \in [\hat{R}_k^{\text{lo}},\,\hat{R}_k^{\text{hi}}]\big]}_{\text{coverage}}
    \cdot
    \underbrace{\max\!\Big(0,\; 1 - \tfrac{\hat{R}_k^{\text{hi}} - \hat{R}_k^{\text{lo}}}{R_k}\Big)}_{\text{tightness}}
    & \text{if interval}, \\[8pt]
    0 & \text{if } \texttt{impossible}.
  \end{cases}
  \label{eq:v2:reward}}
\end{equation}
Coverage is~1 iff the realized remaining cost $R_k$ falls inside the predicted interval. Tightness penalizes wide intervals: a perfect prediction $[R_k, R_k]$ scores~1, while an interval as wide as $R_k$ itself scores~0. For diagnostics, we also report interval hit rate (the fraction of success-case samples covered) and midpoint relative error, $|\frac{\hat{R}_k^{\text{lo}} + \hat{R}_k^{\text{hi}}}{2} - R_k| / R_k$, at the 50th and 90th percentiles.


\vspace{-6pt}
\subsection{Experimental Setup}
\label{sec:v2:method:setup}

\vspace{-6pt}

\textbf{Internal-budget environments.}~~We use
\textbf{Sokoban}~\citep{junghanns1998sokoban}: a planning task where agents push boxes to targets on a $8{\times}8$ grid, capped at $2{,}500$ tokens.
\textbf{Search-R1}~\citep{jin2025searchr1}: multi-hop information retrieval, capped at $3{,}500$ tokens.
\textbf{SWE-bench}~\citep{jimenez2024swebench}: agents resolve GitHub issues, capped at $160$ turns.

\textbf{External-budget environment.}~~
We develop \textbf{Warehouse} (Appendix~\ref{sec:warehouse-environment-details}), a supply-chain environment curated from real enterprise data, with three coupled budget dimensions: cumulative cost (USD), time (weeks), and warehouse occupancy (item-weeks). The agent manages inventory over a 24-week horizon (12 turns), making procurement and allocation decisions against all budgets simultaneously. As the task is continuous cash maximization rather than naturally binary, we evaluate budget awareness via \emph{challenge-conditioned feasibility probes} (App.~\ref{sec:warehouse-budget-construction}), keeping the task success rate balanced 50/50 between reachable and unreachable so that $F_1$ and calibration are identifiable on both sides.

\textbf{Training and Evaluation}~~
We evaluate five frontier models: GPT-5.2 Instant~\citep{openai2026gpt52instant}, Claude Opus 4.7~\citep{anthropic2026adaptiveThinking}, Claude Sonnet 4.6~\citep{anthropic2026sonnet46}, Gemini 3.1 Pro~\citep{google2026gemini31pro}, and Qwen3-235B~\citep{yang2025qwen3}. We use Qwen2.5-7B-Instruct~\citep{qwen2-5} for SFT and RL, and use a combined reward for RL to prevent collapse (Appendix~\ref{app:sft-rl-settings}).


\textbf{Scale.}~~
We generate 128 rollouts per model on Sokoban, Search-R1, and Warehouse, and 64 rollouts on SWE-bench. Each non-terminal turn yields one estimation sample via the rollout-replay protocol, totaling 2{,}000 to 3{,}000 estimation samples per model-task pair.

\vspace{-6pt}
\section{Budget Awareness Decouples from Task Performance}
\label{sec:v2:decoupling}
\vspace{-6pt}

Budget-Aware Agents (BAGEN) obtains distinct capabilities: excelling at completing tasks is not necessarily needed for estimating what those tasks will cost. We show this decoupling holds across all four environments and all three sub-capabilities.

\begin{table}[t]
\centering
\footnotesize
\setlength{\tabcolsep}{3.2pt}
\renewcommand{\arraystretch}{0.92}
\caption{Overall rollout and budget-estimation results. F1@1 and F1@All denote first-turn and all-turn feasibility macro-$F_1$, respectively. Fail-$F_1$ measures detection of \texttt{impossible} cases. Warehouse uses n/a for rollout success because it does not have a separate task-success label.}
\label{tab:v2:overall}

\begin{adjustbox}{max width=\textwidth}
\begin{tabular}{lccccccc}
\toprule
& \multicolumn{2}{c}{\textbf{Task Performance}}
& \multicolumn{3}{c}{\textbf{Feasibility Prediction}}
& \multicolumn{2}{c}{\textbf{Interval Quality}} \\
\cmidrule(lr){2-3}
\cmidrule(lr){4-6}
\cmidrule(lr){7-8}
\textbf{Model}
& \textbf{~~~Success~~~}
& \textbf{~~~Turns~~~}
& \textbf{~~~F1@1~~~}
& \textbf{~~~F1@All~~~}
& \textbf{~~~Fail F1~~~}
& \textbf{~~~Hit~~~}
& \textbf{~~~Reward~~~} \\
\midrule

\rowcolor{blue!4}
\multicolumn{8}{l}{\textbf{SWE-bench}} \\
Claude Opus 4.7
& \textcolor{brown}{\textbf{71.9\%}} & 12.12 & 41.1\% & 51.1\% & 48.8\% & 30.3\% & \textcolor{brown}{\textbf{0.160}} \\
Claude Sonnet 4.6
& 68.8\% & 20.66 & 32.2\% & 37.7\% & 23.4\% & 22.3\% & 0.130 \\
Gemini 3.1 Pro Preview~~~~
& 68.8\% & 37.36 & 39.2\% & \textcolor{brown}{\textbf{58.2\%}} & \textcolor{brown}{\textbf{52.0\%}} & 23.2\% & 0.112 \\
GPT-5.2 Instant
& 57.8\% & 21.52 & 43.5\% & 40.2\% & 21.2\% & \textcolor{brown}{\textbf{44.3\%}} & 0.115 \\
Qwen3 235B
& 33.3\% & 62.92 & \textcolor{brown}{\textbf{47.6\%}} & 35.1\% & 32.8\% & 6.5\% & 0.021 \\

\addlinespace[1pt]
\rowcolor{green!4}
\multicolumn{8}{l}{\textbf{Search-R1}} \\
Claude Opus 4.7
& \textcolor{brown}{\textbf{75.8\%}} & 1.78 & 39.4\% & \textcolor{brown}{\textbf{40.5\%}} & 5.6\% & 23.1\% & 0.114 \\
Claude Sonnet 4.6
& 71.1\% & 1.87 & 37.9\% & 33.3\% & 0.0\% & \textcolor{brown}{\textbf{36.5\%}} & \textcolor{brown}{\textbf{0.154}} \\
GPT-5.2 Instant
& 68.0\% & 3.69 & \textcolor{brown}{\textbf{40.2\%}} & 38.3\% & 0.0\% & 21.4\% & 0.031 \\
Gemini 3.1 Pro Preview~~~~
& 53.9\% & 2.09 & 24.5\% & 24.8\% & 0.0\% & 20.7\% & 0.079 \\
Qwen3 235B
& 35.2\% & 9.94 & 33.2\% & 23.9\% & \textcolor{brown}{\textbf{30.9\%}} & 0.0\% & 0.000 \\

\addlinespace[1pt]
\rowcolor{orange!5}
\multicolumn{8}{l}{\textbf{Sokoban}} \\
Claude Opus 4.7
& \textcolor{brown}{\textbf{56.2\%}} & 5.04 & 46.3\% & 45.6\% & 16.0\% & \textcolor{brown}{\textbf{46.4\%}} & 0.112 \\
Claude Sonnet 4.6
& 51.6\% & 5.65 & \textcolor{brown}{\textbf{46.4\%}} & 53.6\% & 33.9\% & 45.1\% & 0.148 \\
Gemini 3.1 Pro Preview~~~~
& 39.1\% & 5.63 & 40.0\% & \textcolor{brown}{\textbf{61.9\%}} & \textcolor{brown}{\textbf{79.9\%}} & 8.8\% & \textcolor{brown}{\textbf{0.313}} \\
GPT-5.2 Instant
& 35.2\% & 9.02 & 27.7\% & 40.6\% & 32.8\% & 36.0\% & 0.167 \\
Qwen3 235B
& 7.0\% & 10.77 & 6.3\% & 12.6\% & 20.4\% & 10.8\% & 0.029 \\

\addlinespace[1pt]
\rowcolor{purple!4}
\multicolumn{8}{l}{\textbf{Warehouse}} \\
GPT-5.2 Instant
& n/a & 12.00 & 35.0\% & 63.4\% & 56.9\% & 24.7\% & 0.577 \\
Claude Opus 4.7
& n/a & 12.00 & 33.3\% & 63.2\% & 55.7\% & 35.9\% & 0.690 \\
Claude Sonnet 4.6
& n/a & 12.00 & 33.3\% & 64.9\% & 59.0\% & 17.3\% & 0.572 \\
Gemini 3.1 Pro Preview~~~~
& n/a & 12.00 & \textcolor{brown}{\textbf{42.0\%}} & \textcolor{brown}{\textbf{67.0\%}} & \textcolor{brown}{\textbf{62.8\%}} & \textcolor{brown}{\textbf{50.2\%}} & \textcolor{brown}{\textbf{0.698}} \\
Qwen3 235B
& n/a & 12.00 & 41.0\% & 60.8\% & 56.0\% & 17.3\% & 0.483 \\

\bottomrule
\end{tabular}
\end{adjustbox}
\vspace{-8pt}

\end{table}

\vspace{-6pt}

\subsection{Task Success Rate and Budget Estimation is Decoupled}

\vspace{-6pt}

\label{sec:v2:decoupling:actor-vs-estimator}

\textbf{The best actor is not the best estimator.}
On Search-R1, Opus achieves the highest task success rate (75.8\%), yet Sonnet produces better interval estimates (36.5\% hit rate vs.\ 23.1\% for Opus). 
On SWE-bench, the rankings split three ways: Opus leads task success, Gemini leads feasibility prediction, and GPT-5.2 leads interval estimation. 
On Warehouse, rollout success is not reported as a separate outcome, but estimation quality still varies substantially: Gemini achieves the highest interval hit rate (50.2\%), while Sonnet and Qwen are lowest (17.3\%).

\textbf{The correlation is weak.}
Across 20 model-environment pairs, task success rate correlates only weakly with feasibility prediction ($r \approx 0.35$). Left and middle figure of Figure~\ref{fig:v2:reward-vs-rollout-and-progress} visualizes this separation. A model that completes more tasks does not reliably produce better budget estimates. This decoupling suggests that budget awareness draws on different capabilities than task execution, perhaps metacognitive monitoring rather than problem-solving skill.

\begin{figure}[t]
    \centering
    \includegraphics[width=1\linewidth]{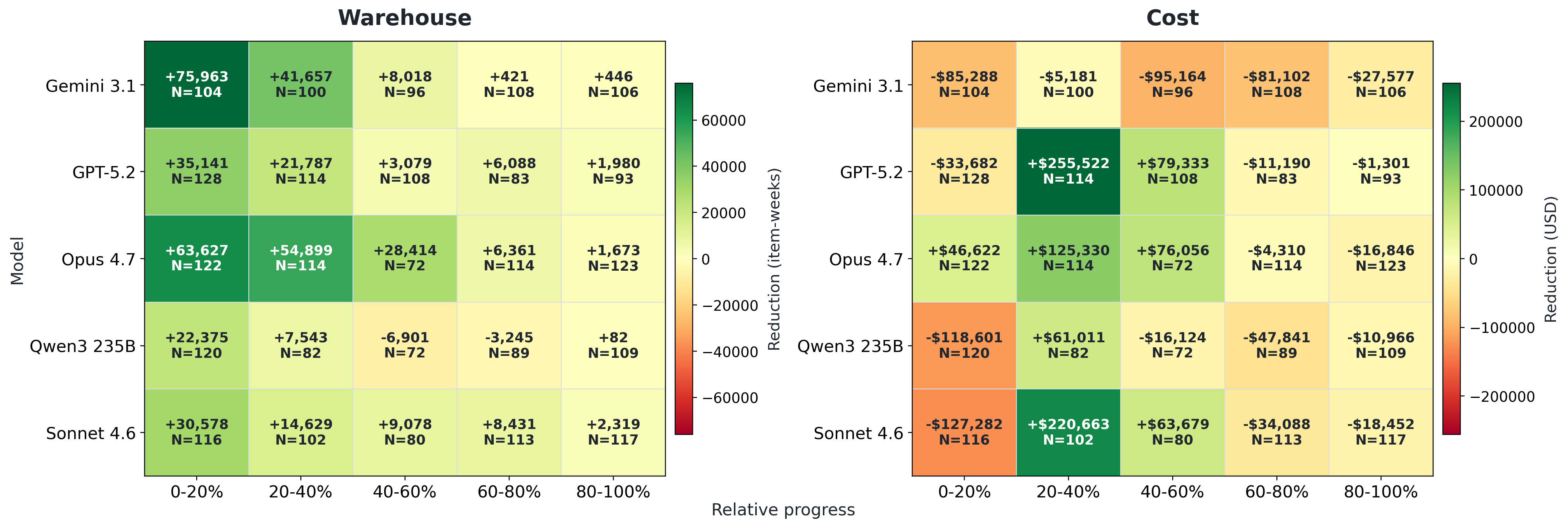}
    \caption{Agent midpoint estimates versus linear extrapolation on Warehouse, where positive cells indicate lower absolute error for the agent than for the extrapolation baseline.}
    \label{fig:agent-advantage-heatmap}
\end{figure}

\textbf{Agent estimates are not just linear extrapolation.}
On Warehouse, we compare each agent midpoint with a deterministic extrapolation baseline, $\widehat{R}_{\mathrm{lin}}=(C_k/k)T-C_k$, and report paired error reduction (Figure~\ref{fig:agent-advantage-heatmap}). Agent predictions reduce error for warehouse occupancy in most model-progress bins, especially early in the rollout. The pattern is mixed for cumulative cost, where extrapolation often remains competitive. Thus, the advantage of budget awareness is real but budget-dimension dependent.

\vspace{-6pt}

\subsection{No Model Dominates All Sub-Capabilities}
\vspace{-6pt}

\label{sec:v2:decoupling:subcapabilities}

We decompose budget awareness into three sub-capabilities: feasibility prediction (binary), early failure detection, and interval calibration. Table~\ref{tab:v2:overall} reports them across all model-environment pairs.

\textbf{No single model leads on all three.}
Gemini achieves the highest binary $F_1$ on Sokoban (61.9\%) but the lowest interval hit rate (8.8\%). On Warehouse, Gemini leads Fail-$F_1$ (62.8\%) yet Qwen produces a low hit rate (17.3\%). Each frontier model has its own profile: a different combination of when it judges feasibility correctly, when it raises infeasibility alarms in time, and how tight its intervals are when the task succeeds.

\textbf{What separates good estimators is calibration, not feasibility prediction.}
Interval hit rate correlates strongly with midpoint bias ($r \approx -0.67$) and width adequacy ($r \approx 0.62$), but only weakly with feasibility $F_1$ ($r \approx 0.35$). Models that correctly predict feasibility often still produce poorly calibrated intervals. Feasibility prediction is necessary but not sufficient for budget awareness, a split that foreshadows the calibration-versus-reasoning analysis in \S\ref{sec:v2:why-fail}.

\vspace{-6pt}
\section{Why Does Budget Estimation Fail?}
\label{sec:v2:why-fail}

\begin{figure}[t]
\centering
\includegraphics[width=0.9\linewidth]{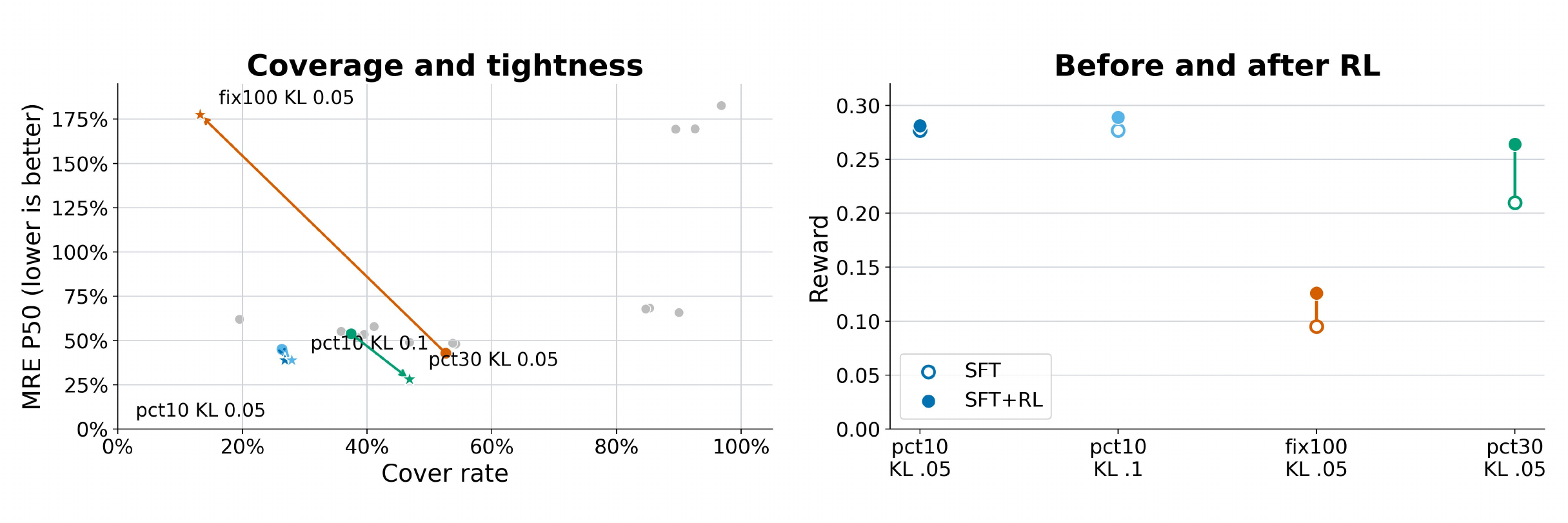}
\caption{Training tradeoffs for interval estimation. Left: coverage versus midpoint error for SFT checkpoints and their SFT+RL continuations. Right: reward before and after RL from the same SFT starts. This indicates that RL can improve estimation performance, but only when it starts from a suitable SFT initialization; without an appropriate SFT warm-start, training collapses.}
\label{fig:v2:sft-rl}
\vspace{-8pt}
\end{figure}
\begin{figure}[t]
    \centering
    \includegraphics[width=\linewidth]{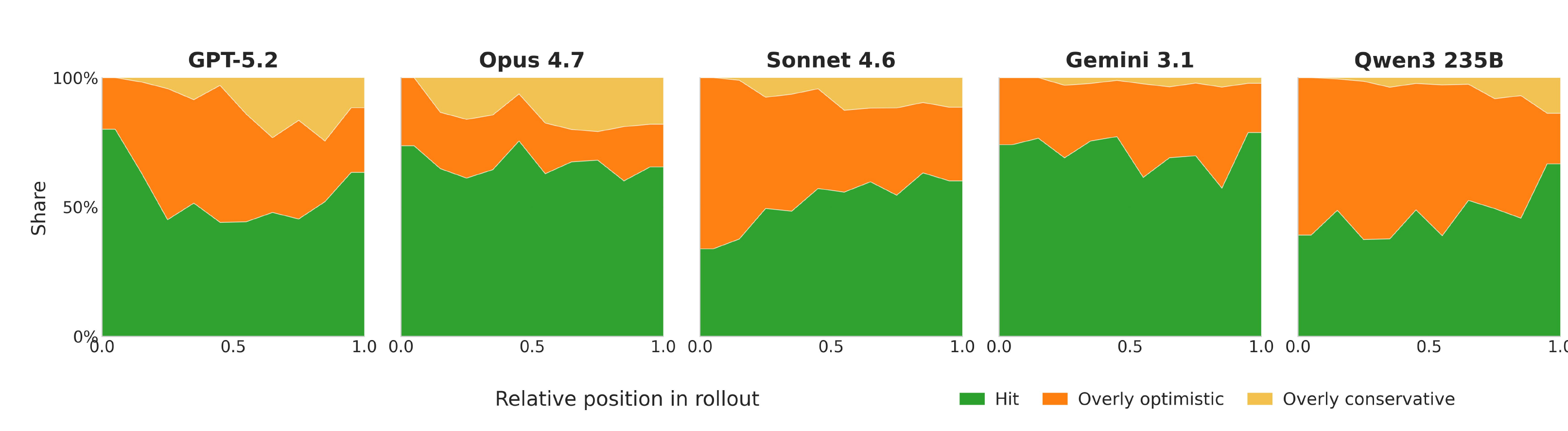}
    \caption{Model generally estimate budget too optimistically throughout the rollout. Conservative bias increases with rollout progress, but remains secondary to optimism overall.}
    \label{fig:v2:optimism-full}
    \vspace{-12pt}

\end{figure}

\vspace{-6pt}

Budget awareness is decoupled from task performance, but we have not yet explained why estimation fails. We test four hypotheses for the underlying mechanism:
\begin{itemize}[leftmargin=*, itemsep=1pt, topsep=1pt]
\item \textbf{Hypothesis 1 (capability gap).} Models lack the reasoning required to predict budget.
\item \textbf{Hypothesis 2 (optimistic prior).} Models have a general bias to underestimate remaining budget.
\item \textbf{Hypothesis 3 (static estimation).} Models cannot leverage execution outcomes to update estimates.
\item \textbf{Hypothesis 4 (late recognition).} Models recognize doomed trajectories only after most of the budget has already been spent.
\end{itemize}

The picture that emerges is mixed. Hypotheses 2 and 4 are supported across all five frontier models. Hypothesis 1 is supported for interval estimation but rejected for binary feasibility, which is a calibration problem. Hypothesis 3 is rejected outright: estimates do update over turns, though the direction of the update is model-specific.

\vspace{-6pt}

\subsection{Binary Feasibility Is Calibration; Interval Estimation Is Reasoning (Hypothesis 1)}
\label{sec:v2:why-fail:h1}

\vspace{-6pt}


We train Qwen-7B on Sokoban budget estimation using supervised fine-tuning (SFT) followed by reinforcement learning (RL). If Hypothesis 1 holds, fine-tuning should leave performance largely unchanged because capability is the bottleneck. If Hypothesis 1 fails, fine-tuning should close the gap, indicating the capability was already there.

\textbf{Binary feasibility is easy to train; precise estimation is hard.}
The base Qwen-7B reaches 25.5\% feasibility prediction accuracy on Sokoban. SFT alone raises accuracy to roughly 90\%, with no RL needed. Feasibility prediction appears to be a calibration problem: the model already has the capability, but needs the right format. Hypothesis 1 is therefore rejected for feasibility prediction.

\textbf{Interval estimation improves more slowly.}
The base model achieves 10.5\% coverage; SFT raises coverage to between 26\% and 53\% depending on target interval width; SFT + RL pushes coverage to 47\% with median midpoint relative error of 28\% (Figure~\ref{fig:v2:sft-rl}). Even after training, nearly half the intervals still miss the true remaining budget. Hypothesis 1 is supported for interval estimation: the bottleneck is reasoning, not output format.

\textbf{RL without SFT collapses.}
RL alone, without an SFT warm-start, fails outright: the model either outputs \texttt{impossible} for everything, or exploits the reward by emitting invalid formats. SFT supplies a format prior that RL cannot recover from sparse reward on its own. This fragility is consistent with Hypothesis 1: when the underlying capability is thin, supervision matters more than reward.

\textbf{Training transfers imperfectly beyond Sokoban.}
Table~\ref{tab:extended-sft-grpo} extends the SFT-then-GRPO pipeline to five budget-estimation settings. Training improves reward in every in-task setting, with the largest gain on Warehouse and the strongest token-budget interval quality on Search-R1. Sokoban and SWE-bench remain harder: models learn feasibility more readily than calibrated intervals. Cross-task evaluation retains only $17$--$36\%$ of in-task reward, suggesting that output format transfers, but budget estimation is largely task-specific.

\begin{table}[t]
\centering
\caption{Extended SFT+GRPO budget-estimator results.  In-task rows compare the untrained estimator reward $R_{\mathrm{base}}$ with the trained reward $R_{\mathrm{train}}$.  Cross-task rows report the source-task trained reward $R_{\mathrm{src}}$, the held-out target-task reward $R_{\mathrm{cross}}$, and reward retention.  Cover is interval coverage on possible cases; MRE$_{50}$ is median midpoint relative error.  A dash indicates that the trained model produced no possible intervals to score.}
\label{tab:extended-sft-grpo}
\small
\setlength{\tabcolsep}{3.5pt}
\resizebox{\textwidth}{!}{%
\begin{tabular}{lllrrrrrrr}
\toprule
Setting & Learner / transfer & Unit & $R_{\mathrm{base/src}}$ & $R_{\mathrm{train/cross}}$ & Gain / retention & Format & Acc. & Cover & MRE$_{50}$ \\
\midrule
\multicolumn{10}{l}{\textbf{In-task training}} \\
Search-R1 & Qwen2.5-7B & Tokens & 0.029 & 0.258 & $+0.229$ & 100.0\% & 56.2\% & 56.7\% & 8.8\% \\
Sokoban & Llama-3.1-8B & Tokens & 0.006 & 0.155 & $+0.149$ & 100.0\% & 78.8\% & 0.0\% & 85.2\% \\
Sokoban & Qwen3-4B & Tokens & 0.002 & 0.143 & $+0.141$ & 88.5\% & 73.0\% & 0.0\% & 2712.5\% \\
SWE-bench & Qwen2.5-7B & Tokens & 0.000 & 0.058 & $+0.058$ & 52.8\% & 28.9\% & 0.0\% & -- \\
Warehouse & Qwen2.5-7B & Cost & 0.004 & 0.555 & $+0.551$ & 100.0\% & 84.6\% & 78.5\% & 16.1\% \\
\addlinespace[2pt]
\multicolumn{10}{l}{\textbf{Cross-task transfer}} \\
Sokoban $\rightarrow$ Search-R1 & Qwen3-4B & Tokens & 0.143 & 0.052 & 36\% & 87.7\% & 40.8\% & 10.5\% & 92.4\% \\
Sokoban $\rightarrow$ Search-R1 & Llama-3.1-8B & Tokens & 0.155 & 0.026 & 17\% & 86.9\% & 44.6\% & 0.0\% & 94.4\% \\
Search-R1 $\rightarrow$ Sokoban & Qwen2.5-7B & Tokens & 0.258 & 0.090 & 35\% & 84.4\% & 50.0\% & 0.0\% & 99.6\% \\
\bottomrule
\end{tabular}%
}
\end{table}

\begin{table}[t]
\centering
\footnotesize
\caption{Aggregate early-stopping tradeoff by estimator model. False-abort rate is the fraction of successful-prefix samples incorrectly labeled \texttt{impossible}. Saved token share is the fraction of failed-rollout tokens saved after the first \texttt{impossible} prediction.}
\label{tab:v2:early-stop}
\begin{tabular}{lrrrr}
\toprule
\textbf{Model} & \textbf{False-abort rate} & \textbf{Saved tokens} & \textbf{False-abort count} & \textbf{Stopped failed rollouts} \\
\midrule
GPT-5.2 Instant & 6.6\% & 64.1\% & 183 / 2{,}776 & 124 / 215 \\
Claude Opus 4.7 & 2.2\% & 28.2\% & 50 / 2{,}234 & 62 / 169 \\
Claude Sonnet 4.6 & 3.3\% & 49.6\% & 76 / 2{,}294 & 101 / 183 \\
Gemini 3.1 Pro & 2.8\% & 55.7\% & 63 / 2{,}266 & 123 / 221 \\
Qwen3 235B & 4.9\% & 38.8\% & 190 / 3{,}909 & 140 / 306 \\
\bottomrule
\end{tabular}
\vspace{-6pt}
\end{table}

\vspace{-8pt}

\subsection{Optimistic Bias Is Universal Across Models and Tasks (Hypothesis 2)}

\vspace{-6pt}
\label{sec:v2:why-fail:h2}

On every progressive interval estimation prediction across Sokoban, Search-R1, SWE-bench, and Warehouse, we record whether the predicted interval misses the realized remaining budget on the optimistic side (predicted budget below realized) or the conservative side (predicted budget above realized). Hypothesis 2 predicts that optimistic misses dominate uniformly.

\textbf{Almost all models underestimate remaining budget.}
Across all 20 model-environment pairs, optimistic misses outnumber conservative ones at every rollout-progress bin (Figure~\ref{fig:v2:optimism-full}). Gemini and Qwen are the most optimistic; Sonnet and Opus are closest to calibrated, but still skew low. The bias is not eliminated by averaging over more turns. Hypothesis 2 is supported.

\textbf{The bias tracks model confidence, not task difficulty.}
Within an environment, weaker models are \emph{more} optimistic about finishing the task, not less. This is the opposite of what would be expected if optimism reflected limited reasoning about hard tasks. The pattern is consistent with overconfidence under limited self-awareness: the model does not know what it does not know.

\vspace{-6pt}

\subsection{Estimates Update Over Turns, but Not Always Toward the Truth (Hypothesis 3)}
\label{sec:v2:why-fail:h3}

\begin{figure}[t]
    \centering
    \includegraphics[width=\linewidth]{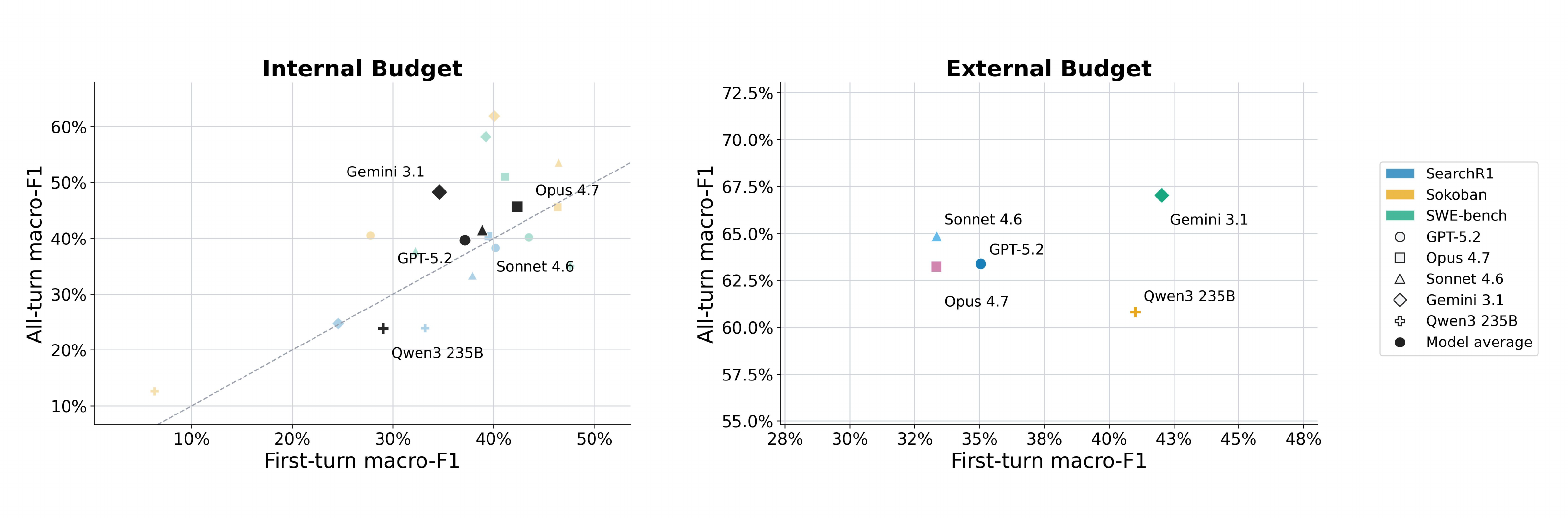}
    \caption{First-turn vs all-turn feasibility $F_1$. Points scatter on both sides of the equality line, so first-turn predictions do not summarize what the same agent would say after seeing partial progress.}
    \label{fig:v2:first-vs-all}
\end{figure}

For each model and environment, we compare feasibility macro-$F_1$ computed using only the first-turn prediction against macro-$F_1$ using all-turn predictions. If Hypothesis 3 holds, the two should be similar because the model cannot use mid-execution evidence. If Hypothesis 3 fails, all-turn $F_1$ should differ from first-turn $F_1$.

\textbf{Later turns produce different estimates than early ones, but not always better.}
On Sokoban, Gemini's macro-$F_1$ improves by $+21.9$ points from first-turn to all-turn evaluation. On SWE-bench, Qwen drops by $-12.5$ points in the opposite direction. Across model-environment pairs, points scatter on both sides of the equality line in Figure~\ref{fig:v2:first-vs-all}. Hypothesis 3 is therefore rejected: estimates do update with execution. The follow-up question is whether updates are refinements toward truth, and the answer is mixed: refinement happens for some model-environment combinations, the opposite for others. The protocol cannot rely on later turns being uniformly better.

\vspace{-6pt}
\subsection{Failure Is Recognized Too Late to Act On (Hypothesis 4)}
\label{sec:v2:why-fail:h4}

\begin{figure}[t]
    \centering
    \includegraphics[width=1\linewidth]{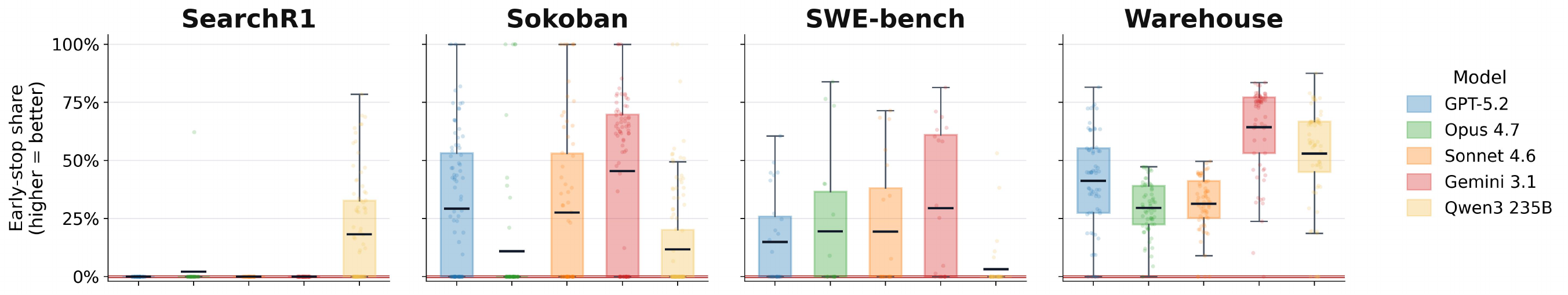}
    \caption{Failure is recognized late: across environments, models often label failed trajectories as \texttt{impossible} only after much of the token budget has already been spent.}
    \label{fig:v2:failure-timing}
\end{figure}

On trajectories that ultimately fail (the agent runs out of budget without solving the task), we measure when within the trajectory the model first predicts \texttt{impossible}. Hypothesis 4 predicts that the alarm fires only late, after most of the budget has been spent. The right panel of Figure~\ref{fig:v2:reward-vs-rollout-and-progress} shows this pattern on Sokoban: failure prediction accuracy improves as the rollout progresses, but the largest gains appear only after substantial budget has already been consumed.

\textbf{Models keep predicting ``feasible'' long after the task is doomed.}
On failed trajectories, models predict feasibility at rates above 70\% even after 60\% of the budget has been consumed. The prediction drops sharply only in the final 20\% of execution, which is too late for meaningful intervention (Figure~\ref{fig:v2:failure-timing}). Hypothesis 4 is supported. \textbf{This late recognition wastes substantial compute}, which we quantify in \S\ref{sec:v2:use:early-stop}.






\vspace{-8pt}

\section{The Signal Is Actionable and Trainable}
\label{sec:v2:use}

\vspace{-8pt}

 The four tests in \S\ref{sec:v2:why-fail} cluster failures into a \emph{calibration regime} (binary feasibility, late recognition) and a \emph{reasoning regime} (precise intervals, persistent optimism). The two admit different
  remedies, which we develop in this section.
  

\vspace{-6pt}
\subsection{Early Stopping Saves Tokens at Low Risk}
\vspace{-6pt}
\label{sec:v2:use:early-stop}

\textbf{A simple policy: stop when the model predicts infeasible.}
At each turn, if the model's estimation outputs \texttt{impossible}, we terminate trajectory. The policy has two error types: a \emph{false abort} stops a trajectory that would have succeeded, and a \emph{false continue} fails to stop one that ultimately fails. False aborts trade success for compute; false continues are missed opportunities to save compute.

\textbf{The savings are substantial with minimal cost.}
Across models, early stopping saves between 28\% and 64\% of tokens on failed trajectories while reducing overall success rate by only 1.6 to 4.2 percentage points (Table~\ref{tab:v2:early-stop}). The estimation signal is already present in the model's predictions; the policy simply acts on it. GPT-5.2 achieves the highest token savings (64\%) but also the highest false-abort rate (6.6\%). Opus is more conservative, saving 28\% of tokens with only 2.2\% false aborts. Per-benchmark behavior is consistent with the aggregate pattern (Table~\ref{tab:v2:early-stop-perbench}): savings are largest on Warehouse and SWE-bench, smaller on Sokoban, and near zero on Search-R1, where rollouts are short enough that infeasibility detection rarely fires before the run ends naturally.

\begin{figure}[t]
\vspace{-6pt}
    \centering
    \includegraphics[width=1\linewidth]{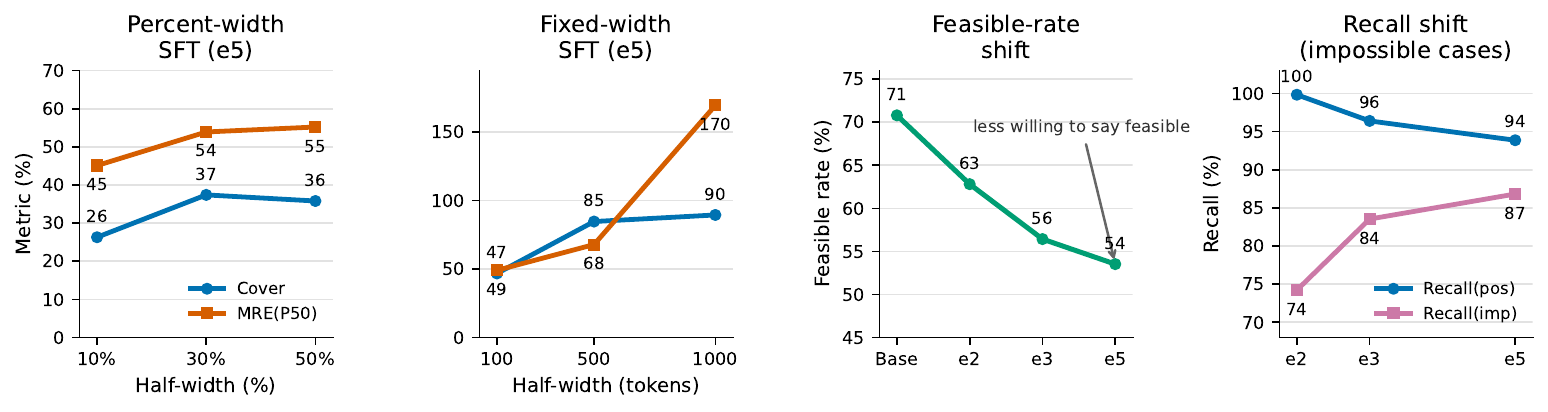}
    \vspace{-6pt}
    \caption{SFT choices control the behavior of the Sokoban budget estimator.
Wider interval targets improve coverage but increase midpoint error, while longer SFT makes the model more conservative by reducing feasible predictions and improving recall on impossible cases.
}
    \label{fig:v2:sft-ablation}
    \vspace{-8pt}
\end{figure}

\vspace{-6pt}
\subsection{Training Dynamics: Width, Epoch, and Initialization}
\vspace{-6pt}
\label{sec:v2:use:training}

Training does more than improve accuracy; it shifts where the model places its confidence. \S\ref{sec:v2:why-fail:h1} reported the headline result that SFT closes most of the binary feasibility gap and SFT followed by RL caps interval coverage near 47\%. Here we characterize three aspects of the training procedure that determine where on the coverage, tightness, and optimism tradeoff curves a trained model lands.

\textbf{SFT interval width controls the coverage-tightness tradeoff.}
Training with narrow target intervals (10\% of remaining budget) produces tight but low-coverage predictions: 26\% coverage with 45\% midpoint relative error. Training with wide intervals (1{,}000 tokens fixed) achieves 90\%+ coverage but poor precision (170\% MRE). The sweet spot is moderate width: 100-token fixed intervals achieve 47\% coverage with 49\% MRE (Figure~\ref{fig:v2:sft-ablation}). Wider targets buy coverage at the cost of precision; narrower targets do the reverse.

\vspace{-6pt}
\section{Related Work}
\vspace{-6pt}
\label{sec:v2:related}

\textbf{Budgets and self-monitoring in LLM agents.}~~ 
A large fraction studies how to \emph{spend} a budget well: constrained and budgeted sequential decision making optimizes reward under resource constraints~\citep{achiam2017constrained,tessler2019reward,carrara2019budgeted,wachi2020safe,zheng2020constrained,liu2022constrainedvariational,li2023nearoptimal,mazumdar2024safe}; budget-aware LLM agents add token,        tool-use, or monetary limits and study cost-aware planning~\citep{liu2026budgetconstrained,ding2026calibratethenact,mccleary2026quantifying}; and adaptive-compute, early-exit, and test-time-scaling methods decide when to stop or skip computation for efficiency~\citep{graves2016adaptive,teerapittayanon2016branchynet,wang2018skipnet,kaya2019shallowdeep,chen2020learningtostop,zeng2023learningtoskip,zhou2026adaptive,snell2025scaling,muennighoff2025s1}. A complementary line asks whether models can assess their own correctness, using reflection, self-feedback, and tool-based critique~\citep{kadavath2022language,xiong2024can,kapoor2024large,yona2024can,deng2024towards,manakul2023selfcheckgpt,madaan2023selfrefine,shinn2023reflexion} and self-correction methods~\citep{gou2024critic,liu2024intrinsicselfcorrection,hu2024uncertainty,wang2024devils}, and analyzes when and why long-horizon agents fail rather than just final outcomes~\citep{wang2024learningfromfailure,barke2026agentrx,yao2023react,yao2022webshop,zhou2023webarena,liu2024agentbench,shridhar2021alfworld,ma2019selfmonitoring,ma2019theregretful}. We ask the dual question: can an agent self-estimate, mid-trajectory, how much budget it will still need? Budget awareness is thus a deployment-relevant slice of self-monitoring where the dominant failure mode is selective over-optimism on failed trajectories \citep{bai2026ai}; we expand this in Appendix~\ref{app:extended-related-work}. 

\textbf{Prediction intervals and calibration.}~~ 
The closest methodological precedent is prediction-interval and calibration work, which studies finite-sample coverage, interval sharpness, conformal validity, and post-hoc calibration~\citep{lei2018distributionfree,pearce2018highquality,kuleshov2018accurate,romano2019conformalized,amini2020deep,chung2021beyond,levi2022evaluating}. These methods assume a fixed predictor on static data; in     our setting the remaining-budget distribution evolves along the trajectory and is partly induced by the agent's own future actions, closer to online belief tracking than offline calibration.

\vspace{-6pt}
\section{Conclusion and Limitations}
\vspace{-6pt}
\label{sec:v2:conclusion}

We introduce Budget-Aware Agents (BAGEN) as a new capability requirement for agents, formalize it as \emph{progressive interval estimation}, and evaluate it on five frontier models across four internal- and external-budget environments via a rollout-replay protocol. The capability decouples from task performance, fails in structured ways (optimistic bias, late failure recognition, calibration-bound feasibility vs.\ reasoning-bound intervals), and is already
actionable through a simple early-stop policy and trainable on Sokoban via SFT-then-RL. Limitations for open directions include extending the training analysis to additional model scales and environments, closing the loop so estimator   
predictions feed back into actor decisions beyond early stopping, supporting fungible budgets across dimensions rather than independent hard constraints, and closing the interval-estimation gap, which our results pinpoint
as the central open problem.



\bibliographystyle{plainnat}
\bibliography{references}

\appendix
\newpage

\section*{Appendix}

\startcontents[appendix]
\printcontents[appendix]{l}{1}{\setcounter{tocdepth}{2}}

\section{Extended Related Work}
\label{app:extended-related-work}
\textbf{Budgeted decision making versus budget estimation.}
Prior work on constrained and budgeted sequential decision making typically formulates budget as an external constraint or environment-provided cost signal, and studies how to maximize reward while satisfying safety, resource, or episode-wise constraints~\citep{achiam2017constrained,tessler2019reward,carrara2019budgeted,wachi2020safe,zheng2020constrained,liu2022constrainedvariational,li2023nearoptimal,mazumdar2024safe}. 
Recent LLM-agent work further introduces explicit resource limits into agentic systems, including token budgets, tool-use limits, monetary API costs, adaptive stopping rules, budget-aware planning, and cost-aware exploration~\citep{zhou2026adaptive,liu2026budgetconstrained,ding2026calibratethenact,mccleary2026quantifying}. 
However, these studies primarily ask how an agent should act when a budget is known, whereas our work asks whether the agent can estimate the remaining budget required to complete a task from an intermediate trajectory state, with uncertainty-aware intervals and infeasibility warnings rather than only realized cost or point forecasts~\citep{romano2019conformalized,kuleshov2018accurate}. 
This distinction is especially important for deployed agents that must manage multiple coupled resources, including model-side internal budgets such as tokens and tool calls and environment-side external budgets such as money, time, inventory, and warehouse capacity. It shows why budget estimation should be evaluated as a standalone capability, rather than inferred from task success or total budget use.~\citep{zhou2026adaptive,liu2026budgetconstrained,ding2026calibratethenact,mccleary2026quantifying}.

\textbf{Prediction intervals, calibration, and uncertainty-aware budget estimation.}
Prediction-interval and calibration work provides the closest methodological precedent because it studies uncertainty-aware prediction: given an input $x$, the predictor outputs an interval $[\hat{y}_{\mathrm{low}}, \hat{y}_{\mathrm{high}}]$ that should contain the true target $y$ with a desired probability, while also remaining as sharp as possible~\citep{lei2018distributionfree,pearce2018highquality,kuleshov2018accurate,romano2019conformalized,amini2020deep,chung2021beyond,levi2022evaluating}. 
Our setting also uses interval-valued predictions. However, the prediction target is fundamentally different. Given a trajectory prefix $\tau_{\leq t}$, the agent must estimate how much additional budget is needed to complete the task from the current state. This remaining requirement depends on partial progress, accumulated mistakes, environment feedback, and the agent's future continuation behavior.
Thus, the target is not a static supervised label attached to an input, but an online, trajectory-dependent quantity that changes as the agent acts; standard calibration assumptions for fixed predictors on static data therefore only partially capture the difficulty of budget estimation~\citep{lei2018distributionfree,romano2019conformalized,levi2022evaluating}. 

Moreover, interval coverage alone is insufficient for deployed agents. 
A remaining-budget interval may cover the realized cost, yet still be operationally poor if it is too wide, systematically optimistic, or fails to warn that the task has become infeasible under the available budget. 
For example, an agent that predicts a broad interval only after exhausting most of its budget is less useful than one that identifies infeasibility early and recommends stopping, requesting more resources, or changing strategy. 
We therefore evaluate uncertainty-aware budget estimation not only through interval hit rate, but also through feasibility prediction and early infeasibility warning, treating budget estimation as an online self-assessment capability rather than a standard offline calibration problem.

\textbf{Adaptive compute as control rather than budget estimation.}
Adaptive-compute and test-time scaling methods are closely related in that they explicitly reason about computational resources, but their objective is usually to learn a compute-control policy rather than to evaluate an agent's self-estimation ability. 
Early-exit, dynamic-routing, learning-to-stop, and test-time scaling methods decide whether to halt, continue, route to a different module, or allocate more inference compute in order to improve the performance--cost tradeoff~\citep{graves2016adaptive,teerapittayanon2016branchynet,wang2018skipnet,kaya2019shallowdeep,chen2020learningtostop,zeng2023learningtoskip,snell2025scaling,muennighoff2025s1}. 
Such decisions can be effective without requiring the model to explicitly state how much additional budget is needed, how uncertain that estimate is, or whether task completion remains feasible under the remaining budget. 
Moreover, this line primarily focuses on internal computation, such as layers, tokens, samples, reasoning steps, or inference-time compute, whereas deployed agents also incur external action-induced costs through tool calls, API usage, elapsed time, monetary spending, inventory changes, or other environment commitments. 
Our setting therefore treats remaining-budget estimation as a distinct capability: the agent must expose an uncertainty-aware belief about future resource requirements across both model-side computation and environment-side action costs, rather than merely choosing a stopping or routing action.

\textbf{Self-monitoring and corrective reasoning.}
A line of work studies whether language models can assess their own correctness, express uncertainty, recognize when they do not know an answer, or detect inconsistencies in generated content~\citep{kadavath2022language,xiong2024can,kapoor2024large,yona2024can,deng2024towards,manakul2023selfcheckgpt,hu2024uncertainty}. 
Related methods use reflection, self-feedback, intrinsic critique, or tool-based criticism to improve model outputs and agent performance over multiple attempts~\citep{madaan2023selfrefine,shinn2023reflexion,gou2024critic,liu2024intrinsicselfcorrection,wang2024devils}. 
These works are closely related because they treat self-assessment as a core model capability, but their focus is usually correctness awareness or corrective reasoning: whether the current answer is likely correct, whether the model should revise it, or how feedback can help produce a better next response. 
Budget awareness asks a different question: even if the agent can judge or improve its answer, can it estimate how much resource is still needed to complete the task, how uncertain that estimate is, and whether completion remains feasible under the available budget?

\textbf{Trajectory-level agent evaluation and budget self-monitoring.}
Interactive agent benchmarks provide the long-horizon, tool-mediated settings in which budget estimation becomes meaningful, since agents must reason, act, replan, and use memory over trajectories whose costs accumulate across turns~\citep{yao2023react,yao2022webshop,zhou2023webarena,liu2024agentbench,shridhar2021alfworld}. 
Recent failure-analysis work further treats failure as a trajectory-level phenomenon, studying root causes, negative rollouts, and where agents fail rather than only final success labels~\citep{wang2024learningfromfailure,barke2026agentrx}. 
Closest in spirit, progress-monitoring methods use auxiliary estimates of task progress to guide search, backtracking, and action choice~\citep{ma2019selfmonitoring,ma2019theregretful}. 
Our focus is different from both retrospective failure diagnosis and progress estimation: given a partial trajectory, we ask whether the agent can estimate the remaining token or financial budget, express uncertainty, and warn that completion may be infeasible under the available resources. 
This exposes selective over-optimism on failed rollouts, a failure mode not captured by standard success, path-efficiency, or progress metrics.

\begin{figure}[t]
\centering
\scriptsize
\setlength{\tabcolsep}{3pt}
\renewcommand{\arraystretch}{1.1}
\resizebox{\textwidth}{!}{%
\begin{tabular}{|p{0.22\textwidth}|c|c|c|c|c|c|c|c|}
\hline
Similar Work &
\begin{tabular}{@{}c@{}}Agentic\\setting\end{tabular} &
\begin{tabular}{@{}c@{}}Trajectory-\\level\end{tabular} &
\begin{tabular}{@{}c@{}}Online during\\execution\end{tabular} &
\begin{tabular}{@{}c@{}}Interval /\\calibrated UQ\end{tabular} &
\begin{tabular}{@{}c@{}}Token\\budget\end{tabular} &
\begin{tabular}{@{}c@{}}Financial\\budget\end{tabular} &
\begin{tabular}{@{}c@{}}Endogenous\\future cost\end{tabular} &
\begin{tabular}{@{}c@{}}Failure-specific\\analysis\end{tabular} \\
\hline
Can LLMs Express Their Uncertainty?~\citep{xiong2024can} & \xmark & \xmark & \xmark & \pmark & \xmark & \xmark & \xmark & \pmark \\
\hline
Conformalized Quantile Regression~\citep{romano2019conformalized} & \xmark & \xmark & \xmark & \cmark & \xmark & \xmark & \xmark & \xmark \\
\hline
Adaptive Stopping for Multi-Turn LLM Reasoning~\citep{zhou2026adaptive} & \pmark & \cmark & \cmark & \cmark & \pmark & \xmark & \pmark & \xmark \\
\hline
Budget-Constrained Agentic LLMs (INTENT)~\citep{liu2026budgetconstrained} & \cmark & \cmark & \cmark & \xmark & \pmark & \cmark & \cmark & \xmark \\
\hline
Calibrate-Then-Act~\citep{ding2026calibratethenact} & \cmark & \cmark & \cmark & \xmark & \xmark & \xmark & \pmark & \xmark \\
\hline
Budget-Constrained Agentic Search (BCAS)~\citep{mccleary2026quantifying} & \cmark & \cmark & \cmark & \xmark & \cmark & \xmark & \pmark & \xmark \\
\hline
Our work & \cmark & \cmark & \cmark & \cmark & \cmark & \cmark & \cmark & \cmark \\
\hline
\end{tabular}}
\caption{Delta map between our formulation and the closest prior work. Rows are representative neighboring papers; columns are the main components of the problem definition. \cmark{} means the component is directly modeled, \pmark{} means partially related, and \xmark{} means largely absent.}
\label{fig:related-work-delta-map}
\end{figure}

The central gap highlighted by Figure~\ref{fig:related-work-delta-map} is that prior work typically covers only one or two sides of the space at a time. Our work is positioned at the intersection of these dimensions and formulates online remaining-budget interval estimation as a distinct capability of agents.

\section{Detailed Experimental Settings}
\subsection{Environments and Tasks}
\label{sec:environments-and-tasks}

Our benchmark spans four environments, SearchR1, Sokoban, and SWE-bench use token budgets; Warehouse uses a joint financial budget(cost/time/inventory) over multiple resources.

\textbf{Sokoban.}
We use a Sokoban environment to study budget estimation under irreversible planning. The rollout agent solves procedurally generated $8 \times 8$ puzzles with two boxes and search depth 30 under a 2{,}500-token budget. For rollout generation, we cap each response at 800 tokens and truncate dialogue history at 2{,}500 context tokens.

\textbf{SearchR1.}
We use the SearchR1 environment backed by a HotpotQA-derived parquet dataset and a retrieval server. At each turn, the agent must either issue one search query or submit a final answer under a 3{,}500-token budget. 

\textbf{SWE-bench}
SWE-bench evaluates coding agents on realistic GitHub issue-resolution tasks. We include it because coding agents are common in practice, and coding tasks are token-intensive. Their token usage often comes from repeated repository inspection, targeted edits, test execution, error analysis, and repair loops. This makes SWE-bench a natural benchmark for evaluating remaining-budget estimation.

\textbf{Warehouse.}
We use a warehouse-management control environment with coupled resource constraints. At each step, the agent must allocate inventory, financing, and cash-flow decisions so the business can continue toward a profitable outcome without violating limits. This benchmark tests a different notion of budget awareness: instead of internal reasoning cost, the agent must estimate whether it can still finish while respecting external operational budgets. Detailed budget construction appears in Section~\ref{sec:warehouse-budget-construction}.

\subsection{Evaluation Protocol}
\label{sec:evaluation-protocol}

Estimator and rollout roles are separated in all experiments. Each rollout is first generated and logged by a rollout model, and the estimator is then evaluated on prefixes of that logged trajectory. In the main experiments, the estimator uses the same model as the rollout generator, corresponding to self-estimation. 

At each evaluation point, we replay the rollout prefix as dialogue history and ask the estimator to predict the budget needed from the next turn onward. The estimator only sees information available up to the current turn, including the replayed history, the budget limits, and summaries of completed turns. For token-budget tasks, these summaries include per-turn token usage; for Warehouse, they include cumulative progress and resource usage so far. The estimator does not receive future turns, future tool outputs, terminal success labels, or the realized remaining budget. We also exclude the final terminal step from estimation, since there is no future budget to predict after the trajectory has ended.

The estimator must output either an interval \verb|<answer>[est_low, est_high]</answer>| for the remaining total token usage from the next turn onward, or \verb|<answer>impossible</answer>| if it predicts that the trajectory can no longer finish successfully within the budget. For Warehouse, the estimator receives the target cash threshold and three resource budgets, \texttt{time\_weeks}, \texttt{warehouse\_item\_weeks}, and \texttt{cumulative\_cost\_usd}, together with cumulative usage so far. It must predict whether the trajectory can still reach the target cash while satisfying all three constraints; if feasible, it outputs one interval for each remaining resource, otherwise it outputs impossible. 

Each non-terminal rollout prefix yields one evaluation sample. If a trajectory has $T$ turns, it contributes $T-1$ samples, because the final turn is not estimated. This construction preserves the online estimation setting: the estimator always predicts from a partial trajectory using only information available at that point. 

SWE-bench requires additional subsampling because coding-agent trajectories are much longer than those in the other benchmarks. 
In the current SWE-bench runs, this produces 712--3{,}715 candidate prefixes per rollout family, so using all prefixes would make long coding trajectories dominate the evaluation set. 
We therefore cap each SWE-bench rollout family at 512 estimation prefixes. 
To construct this split, we sort rollouts by their number of assistant turns, partition them into eight nearly equal-sized length buckets, and sample prefixes with a seeded randomized round-robin strategy across buckets and rollouts using random seed 42. 
This procedure buckets at the rollout level rather than the prefix level, so longer rollouts do not receive proportionally more sampling mass simply because they contain more prefixes. This makes the split fair because it balances coverage across rollout-length regimes and rollout instances, preventing unusually long trajectories from contributing a disproportionate number of highly correlated prefixes while keeping the SWE-bench evaluation size comparable to the other benchmarks.

\subsection{Pilot Study Setup}                                                                                                                                                                                               
\label{app:pilot-setup}                                                                                                                                                                                                                                                                                                                                                                                                                                       The pilot study in \S\ref{sec:v2:naive-probe} establishes the two failures of single-point budget estimation that motivate progressive interval estimation. We describe its setup here.                                        
\textbf{Tasks.}                                                                                                                                                                                                           
We use two internal-budget environments. \textbf{Sokoban}~\citep{junghanns1998sokoban} is a planning task on an $8{\times}8$ grid with a $2{,}500$-token cap; it admits a clean ground-truth optimal budget per puzzle.
\textbf{Search-R1}~\citep{jin2025searchr1} is a multi-hop information-retrieval task with a $3{,}500$-token cap; the realized budget is variable and driven by retrieval depth. The two tasks are deliberately different in    shape, so a failure showing up on both is unlikely to be an artifact of one task's structure.
\textbf{Elicitation.}                                                For each rollout we elicit two estimates from the same model.
A \emph{first-turn estimate} is taken at task start, before any action, asking the model to predict the total tokens it will spend on the task; this matches the single-point prompting protocol used in prior work.         
A \emph{later-turn estimate} is taken after replaying the logged prefix $\tau_{\leq k}$ of the same rollout, asking the model to predict the \emph{remaining} tokens from turn $k{+}1$ onward. The later-turn estimate is    
therefore a sequence of single-point predictions, one per non-terminal turn, made by the same model that produced the rollout.                                                                                                                                                                                                                                                                                                                              
\textbf{Models and scale.}                                                                                                                                                                                                
We evaluate the same five frontier models as in the main paper: Claude Opus 4.7~\citep{anthropic2026adaptiveThinking}, Claude Sonnet 4.6~\citep{anthropic2026sonnet46}, GPT-5.2 Instant~\citep{openai2026gpt52instant},
Gemini 3.1 Pro~\citep{google2026gemini31pro}, and Qwen3-235B~\citep{yang2025qwen3}. We sample $128$ rollouts per (model, task) pair.                                                                                           
\textbf{Scoring.}                                                                                                                                                                                                         
\emph{Optimism analysis} (Figure~\ref{fig:v2:optimism-full}, first failure): for each prediction $\hat{B}$ we compare it to the realized rollout cost $B$ and classify it as \emph{optimistic} ($\hat{B} < B$) or
\emph{conservative} ($\hat{B} > B$); we report the share of each across rollout-progress bins.                                                                                                                                 \emph{First-turn vs.\ later-turn comparison} (Figure~\ref{fig:v2:first-vs-all}, second failure): we collapse each single-point prediction into a binary feasibility label by checking whether $\hat{B}$ stays under the
budget cap, and compute feasibility macro-$F_1$ over either first-turn predictions only or all per-turn predictions (definition in Eq.~\ref{eq:v2:macro-f1}). The two are then plotted against each other on equal axes.

\subsection{Warehouse Budget Construction}
\label{sec:warehouse-budget-construction}

Warehouse is evaluated as a multi-resource financial estimation task that models a manufacturing firm making weekly operational decisions, such as producing goods, replenishing inventory, drawing supplier credit, repaying debt, and factoring accounts receivable to improve cash flow. Full details on the underlying data source, design goals, calibration choices, and reward structure are deferred to Appendix~\ref{sec:warehouse-environment-details}. For each rollout prefix, the estimator receives the dialogue history, current cash, completed-week summaries, cumulative resource usage, and per-step resource consumption so far. It must predict whether the rollout can still reach the target final cash threshold while staying within all resource budgets. If feasible, it outputs one interval for each remaining resource; otherwise, it outputs \texttt{impossible}. This setting creates coupled trade-offs: increasing inventory may improve sales but raises warehouse occupancy, drawing credit may improve short-term cash but adds repayment pressure, and delaying production may reduce costs but lower final cash. Thus, Warehouse tests whether estimators can reason about realistic multi-resource constraints rather than a single token budget.

Warehouse contains three budget dimensions. The first is \texttt{time\_weeks}, which limits the total planning horizon. The second is \texttt{warehouse\_item\_weeks}, which measures cumulative warehouse occupancy over time, i.e., inventory integrated across weeks. The third is \texttt{cumulative\_cost\_usd}, which limits total operational spending.

\textbf{Why we construct feasibility probes.}
Warehouse is a continuous-optimization task (more cash is better, fewer resources used is better) and therefore has no natural binary success label. To evaluate budget awareness in this environment, we construct \emph{challenge-conditioned feasibility probes}: each instance pairs a logged rollout with a sampled (target cash, time, warehouse, cost) tuple, and feasibility is defined as whether that rollout still satisfies the target and all three resource constraints. We balance reachable and unreachable probes 50/50 so that macro-$F_1$, Fail-$F_1$, and calibration are identifiable on both sides; under a heavily skewed split (e.g., 90/10), a model that always predicts ``feasible'' would score deceptively well. This 50/50 balance is an evaluation-design choice, not a claim about deployment prevalence: we do not assert that real supply-chain decisions fail half the time, only that controlled coverage of clearly-feasible, borderline-feasible, borderline-infeasible, and clearly-infeasible regions is needed to test whether models know their own state.

\textbf{Sampling procedure.}
The estimator uses the \texttt{half\_reachable} budget preset with random seed 42.
We first shuffle rollouts and assign half to a reachable group and half to an unreachable group. 
For reachable rollouts, the target cash is sampled from $U(0.50,1.00)\times$ final cash, clipped so that it does not exceed final cash; the time budget is set to the realized trajectory length, and the warehouse and cost budgets are sampled uniformly between $1.0\times$ and $1.2\times$ their realized final totals. 
For unreachable rollouts, the target-cash, time-budget, warehouse-budget, and cost-budget ratios are independently sampled from $U(0.50,2.00)$ until the logged rollout fails the feasibility check. 
Thus unreachable cases may be caused by an overly high target cash, too-tight resource budgets, or a combination of both.

\subsection{Experiment Matrix}
\label{sec:experiment-matrix}

The experiment matrix is shown in Table~\ref{tab:experiment-matrix}.

\begin{table*}[t]
\centering
\small
\setlength{\tabcolsep}{4pt}
\renewcommand{\arraystretch}{1.15}
\begin{tabularx}{\textwidth}{l l c X X}
\toprule
Benchmark & Budget Type & Rollouts & Horizon & Estimation Target \\
\midrule
SearchR1 & Token & 128 &
Up to 3500 tokens; one search/action per turn &
Remaining tokens within 3{,}500. \\

Sokoban & Token & 128 &
Up to 2500 tokens; up to 3 actions per turn &
Remaining tokens within 2{,}500. \\

SWE-bench & Token & 64 & up to 160 turns
&
Remaining tokens under rollout-family-specific median budgets.\\

Warehouse & Financial & 128 &
2 weeks per turn; totally 11 turns &
Remaining time, warehouse item-weeks, and cost while reaching target cash. \\
\bottomrule
\end{tabularx}
\caption{Experiment matrix for the four evaluation benchmarks. Each row states the budget modality, rollout count, horizon constraint, and remaining-budget target predicted from each logged prefix.}
\label{tab:experiment-matrix}
\end{table*}

\subsection{Model Identifiers and Inference Settings}
\label{sec:model-identifiers-and-inference-settings}

To ensure reproducibility, we report exact deployment metadata for every model used in rollout generation and estimation. The display names in the main text are shorthand; the appendix table gives the exact API model identifier and invocation configuration.

For each model, we log display name, provider, exact API model id or local checkpoint hash, date queried, endpoint or region, reasoning configuration, max output tokens, temperature, top-$p$ if used, stop settings, and sampling-seed policy.

The term ``Low Thinking'' denotes a constrained provider-native reasoning configuration. Depending on the backend, this corresponds to one of: \texttt{reasoning\_effort=low} (OpenAI), \texttt{output\_config.effort=low} (Anthropic), or \texttt{thinking\_mode=low} (Google/Qwen). The appendix table reports the exact control for each model.

Table~\ref{tab:model-identifiers} lists the concrete entries used in this paper.

\begin{sidewaystable*}[t]
\centering
\scriptsize
\setlength{\tabcolsep}{3pt}
\renewcommand{\arraystretch}{1.15}
\begin{tabularx}{\textwidth}{l l l X l c c}
\toprule
Display Name & YAML Key & Provider & Model ID & Reasoning & Max Tokens & Temp. \\
\midrule
GPT-5.2 Instant &
OpenAI-5.2-Instant &
openai &
\texttt{gpt-5.2} &
\texttt{reasoning\_effort: none} &
800 &
N/A \\

Claude Opus 4.7 Low Thinking &
Claude-Opus-4.7-low-thinking &
anthropic &
\texttt{claude-opus-4-7} &
{\small\texttt{output\_config.effort: low}} &
800 &
N/A \\

Claude Sonnet 4.6 Low Thinking &
Claude-Sonnet-4.6-low-thinking &
anthropic &
\texttt{claude-sonnet-4-6} &
{\small\texttt{output\_config.effort: low}} &
800 &
N/A \\

Gemini 3.1 Pro Preview &
OpenRouter-Gemini-3.1-Pro-Preview &
openrouter &
\texttt{google/gemini-3.1-pro-preview} &
\texttt{thinking\_mode: low} &
800 &
0 \\

Qwen3 235B &
qwen/qwen3-235b-a22b-2507 &
openrouter &
\texttt{qwen/qwen3-235b-a22b-2507} &
\texttt{thinking\_mode: low} &
800 &
0 \\
\bottomrule
\end{tabularx}
\caption{Model deployment identifiers and inference settings used for rollout generation and budget estimation. The table maps each display name to its provider, exact model ID, reasoning configuration, maximum output length, and temperature.}
\label{tab:model-identifiers}
\end{sidewaystable*}

\section{Warehouse Environment Details}
\label{sec:warehouse-environment-details}

\begin{figure*}[t]
\centering
\includegraphics[width=\linewidth]{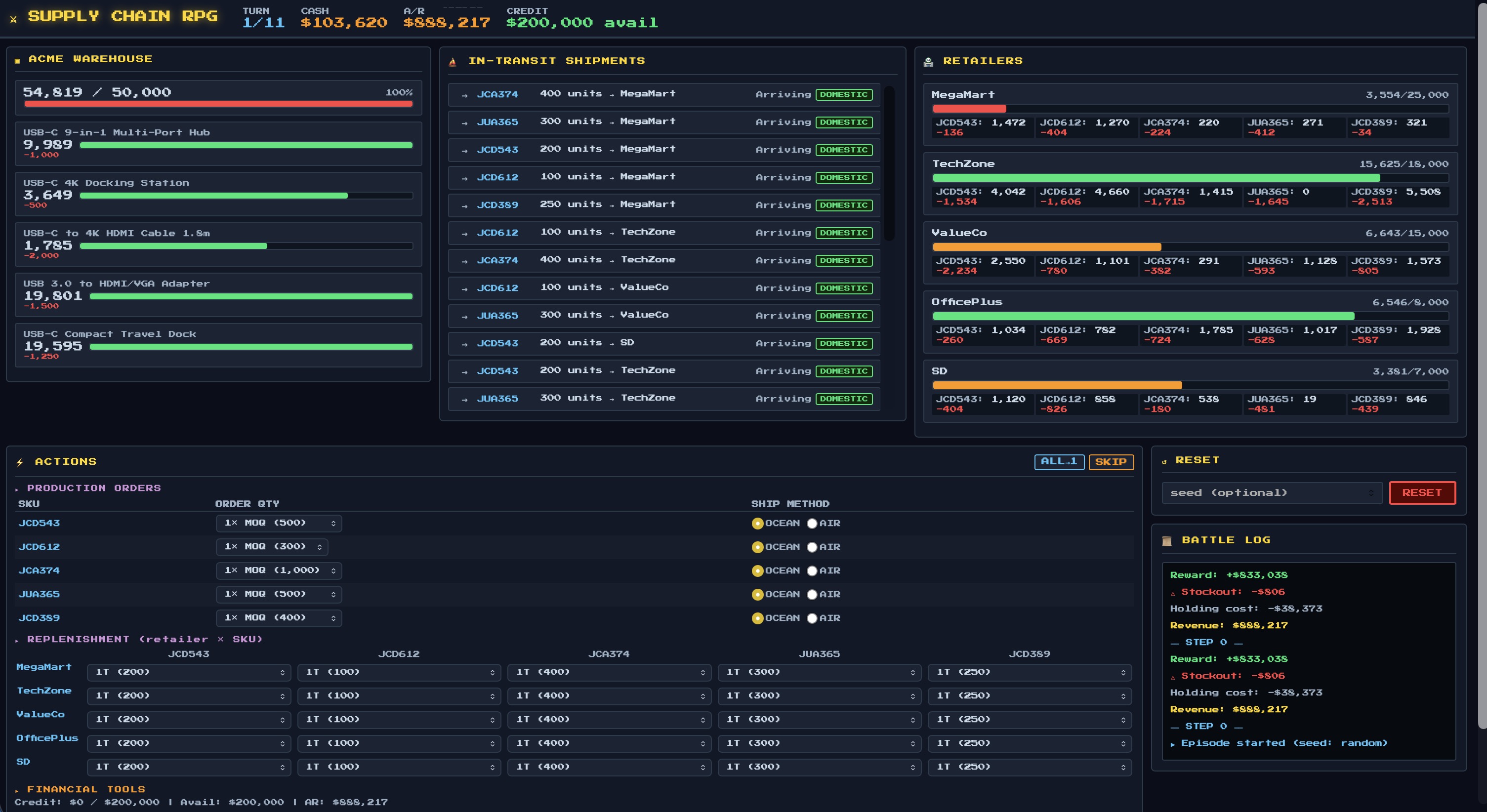}
\caption{Overview of the Warehouse environment. The agent operates a manufacturing firm that orders from OEMs, ships internationally to its own warehouse, and fulfills weekly demand from five retailer accounts under joint money/space/time constraints.}
\label{fig:warehouse-environment}
\end{figure*}

\subsection{Data Source}
\label{sec:warehouse-data-source}

The demand panel is sourced from a real mid-sized US consumer-electronics distributor (anonymized as \textit{Acme}) and its five downstream retail accounts, collapsed into archetypes (\textit{MegaMart}, \textit{TechZone}, \textit{ValueCo}, \textit{OfficePlus}, \textit{SupplyDirect}) covering mass discount, electronics specialty, value/household, office/B2B, and small-account direct channels. We obtained \textbf{22 consecutive weeks} of unit-level weekly sell-through per (retailer, SKU) pair across five high-volume product families (USB-C hub, 4K dock, USB-C/HDMI cable, USB display adapter, travel dock). Manufacturers and order quantities are \textit{not} in the dataset---those are agent decisions in our environment. Wholesale prices, retailer warehouse capacities, payment terms (Net-30 across the board), and stockout/overstock chargeback rates were taken from retail vendor agreements with light rounding; manufacturing costs, MOQs, production lead times, and unit weights were chosen to reflect typical consumer-electronics OEM contract terms. We did not synthesize or smooth the demand series---week-to-week variance, step-changes, and the occasional collapsing series (e.g., one \textit{OfficePlus} SKU) are present in the raw data.

At each episode reset the five demand series belonging to a retailer are randomly permuted across the five game SKUs, so SKU identities do not inherit fixed demand profiles and the agent must read demand from the observation rather than from any learned prior over individual SKUs. Per-step base demand is the sum of two consecutive weeks of the assigned series with additive Gaussian noise scaled by the series standard deviation. We do not layer additional seasonality on top---the 22-week window already captures it.

\subsection{Design Goal: Stressing Money, Space, and Time}
\label{sec:warehouse-design-goal}

A competent policy must trade off three orthogonal pressures at once: \textbf{money} (tight \$500K initial cash, 30/70 split mfg payment, Net-30 receivables, holding + OpEx burn, a \$200K credit line at 0.8\% per step, and 5\% AR factoring), \textbf{space} (50,000-unit Acme warehouse plus per-retailer DCs of 7K--25K units, with overstock chargebacks above 95\% occupancy), and \textbf{time} (production lead 25--45 days, ocean vs.\ air international transit at 32 vs.\ 6 days, domestic 3--4 days, Net-30 cash collection). Every mechanism is anchored to at least one of these axes.

\subsection{Calibration Decisions}
\label{sec:warehouse-calibration}

Two parameter choices were materially tuned away from naive defaults during development; the rest follow from real-world anchoring.

\textbf{Initial inventory is set to 0.}
Earlier calibrations pre-stocked Acme and the retailers with several weeks of expected demand. Trajectories were then largely shaped by the starting conditions rather than by the policy: pre-stocked inventory absorbed the early turns of demand regardless of what the agent did, and roughly 3 of 11 turns ran on autopilot before policy choices began to matter. We now start every (retailer, SKU) pair and the Acme warehouse at \textbf{zero units} so that outcomes are driven by model decisions end-to-end and the early-episode cash crunch (no AR matured yet, no inventory to ship) is genuine.

\textbf{International transit and production lead times are compressed.}
Real Shenzhen$\rightarrow$US-West-Coast door-to-door shipping is 45--60 days ocean and 7--10 days air; OEM production for the more complex SKUs realistically runs 60--90 days once material procurement is included. Plugged in directly, a single ocean order placed at $t = 0$ would arrive around day $90 + 60 = 150$---turn 11 of an 11-turn episode at our 14-day step granularity, meaning 4--5 turns elapse before any production decision surfaces. We instead use ocean $= 32 \pm 4$ days, air $= 6 \pm 1$ days, and production lead $= 25$--$45$ days by SKU. The fastest end-to-end loop (cable, air) then lands in $\approx 2$ steps and the slowest (dock, ocean) in $\approx 5.5$ steps---short enough that the agent observes consequences within the episode, long enough that time pressure is load-bearing.

\textbf{Other calibration choices, briefly.}
Initial cash is \$500K (not \$2M) so the working-capital decision actually binds---at $\sim$\$116K weekly GMV and Net-30 terms, $\sim$\$464K of cash is tied up in any one AR cycle. The mfg payment is a \textbf{30/70 split} (deposit at order, balance on completion) rather than 100\%-on-order, so cash flow is sensitive to the projected trajectory across the production horizon and orders can stall in queue if cash is short on completion. Production quantities are quantized to \textbf{1$\times$/2$\times$/3$\times$ MOQ} with 0\% / 8\% / 15\% volume discounts, mirroring how factories actually quote. Episode length is \textbf{22 weeks $\div$ 14 days per step $=$ 11 steps}---short enough that endgame myopia is diagnosable, long enough for two full ocean-production cycles. Bankruptcy is a \textbf{two-stage floor} (\$50K per-step penalty whenever cash $< 0$, hard truncation at cash $< -\$200$K) rather than a single hard rule, so a small early-turn underprediction does not end the episode prematurely.

\subsection{Reward and Termination}
\label{sec:warehouse-reward}

The per-step reward is
\begin{equation}
r_t \;=\; \pi_t \;-\; p^{\text{stockout}}_t \;-\; p^{\text{overstock}}_t \;-\; 50{,}000 \cdot \mathbf{1}[\text{cash}_t < 0],
\end{equation}
where $\pi_t$ is within-step operating profit (revenue minus holding cost minus fixed OpEx), and $p^{\text{stockout}}_t$, $p^{\text{overstock}}_t$ are the chargeback penalties from unfilled retailer demand and from $>$95\%-occupied retailer DCs. Manufacturing costs, international and domestic shipping costs, credit interest, and AR-factoring fees affect cash but \textbf{do not} enter the reward---they are investments and financing costs whose payoff arrives, if at all, through future operating profit. This is the central credit-assignment challenge: a production decision at $t$ pays back through revenue at $t+4$ to $t+6$. Including these costs in the reward collapses the environment to a near-myopic control problem and removes precisely the long-horizon coordination that motivated the design. The episode terminates at $t = T$ (default 11) and is force-truncated if cash drops below $-\$200$K.

\subsection{Theoretical Maximum Reward (Defined but Unused)}
\label{sec:warehouse-theoretical-max}

For completeness we derive a hypothetical ceiling on cumulative reward under an idealized policy that (i) realizes mean demand each step with no noise, (ii) keeps Acme inventory exactly equal to the step's total demand so holding cost is minimal but no stockouts occur, and (iii) faces no cash constraint. Under those assumptions, period revenue, holding cost, and OpEx all become deterministic, and the cumulative ceiling is
\begin{equation}
R_{\max} \;=\; \sum_{t=1}^{T} \left[ \sum_{r,s} d_{r,s} \cdot p_{r,s} \cdot w \;-\; h \cdot I_t \cdot w \;-\; \text{OpEx} \cdot w \right],
\end{equation}
where $d_{r,s}$ is mean weekly demand of SKU $s$ at retailer $r$, $p_{r,s}$ is the wholesale price, $w$ is the number of weeks per step (default 2), $h = \$0.35$/unit/week is the holding rate, $I_t = \sum_{s} d_s \cdot w$ is the per-step ideal Acme warehouse level (just enough to cover the step's demand), and OpEx $= \$8{,}000$/week is the fixed operating cost. A ``combined score'' variant adds back ending cash and the manufacturing-cost value of ending inventory at both Acme and the retailers, treating leftover stock as recoverable capital.

\section{Additional Experimental Visualizations}

\begin{figure}[htbp]
\centering
\includegraphics[width=\linewidth]{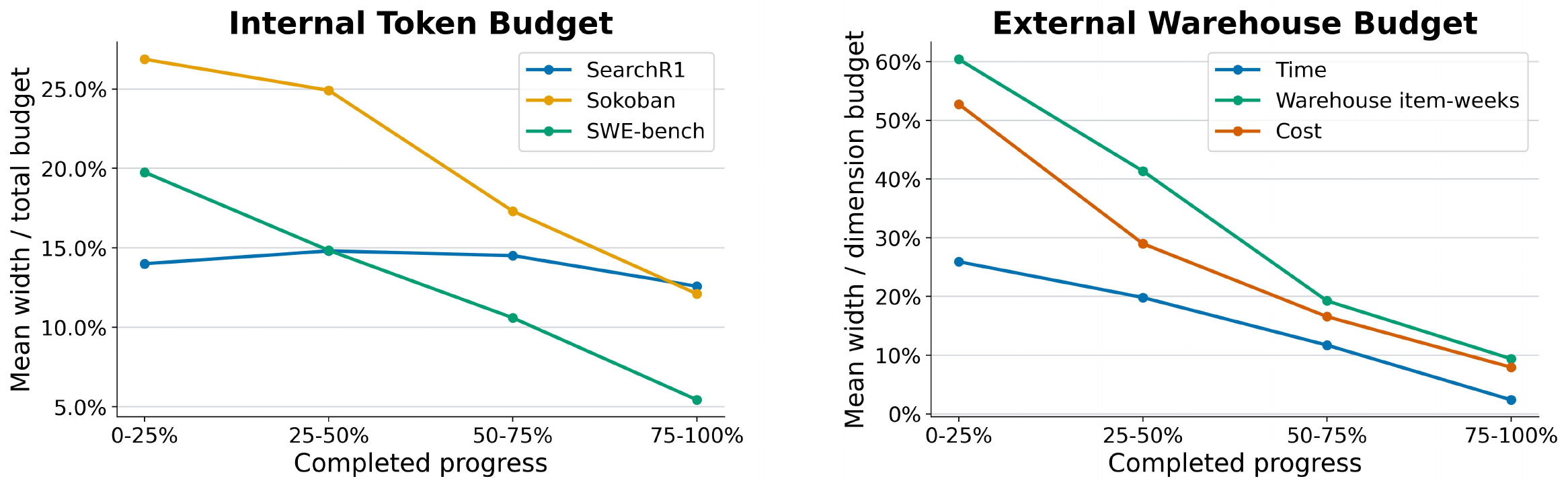}
\caption{Predicted interval width changes over task progress. Internal-task widths are normalized by token budget; Warehouse reports separate normalized widths for time, warehouse item-weeks, and cost.}
\label{fig:appendix-interval-width-progress}
\end{figure}

This appendix presents analysis for budget-estimation behavior.
Figure~\ref{fig:appendix-interval-width-progress} separates the internal token-budget setting from the external Warehouse setting.
For internal tasks, interval width is normalized by the total token budget.
For Warehouse, we report the same quantity for time, warehouse item-weeks, and cost.
This pattern suggests that the estimators partly learn to update their uncertainty as more rollout evidence becomes available.

\begin{figure}[t]
    \centering
    \includegraphics[width=0.95\linewidth]{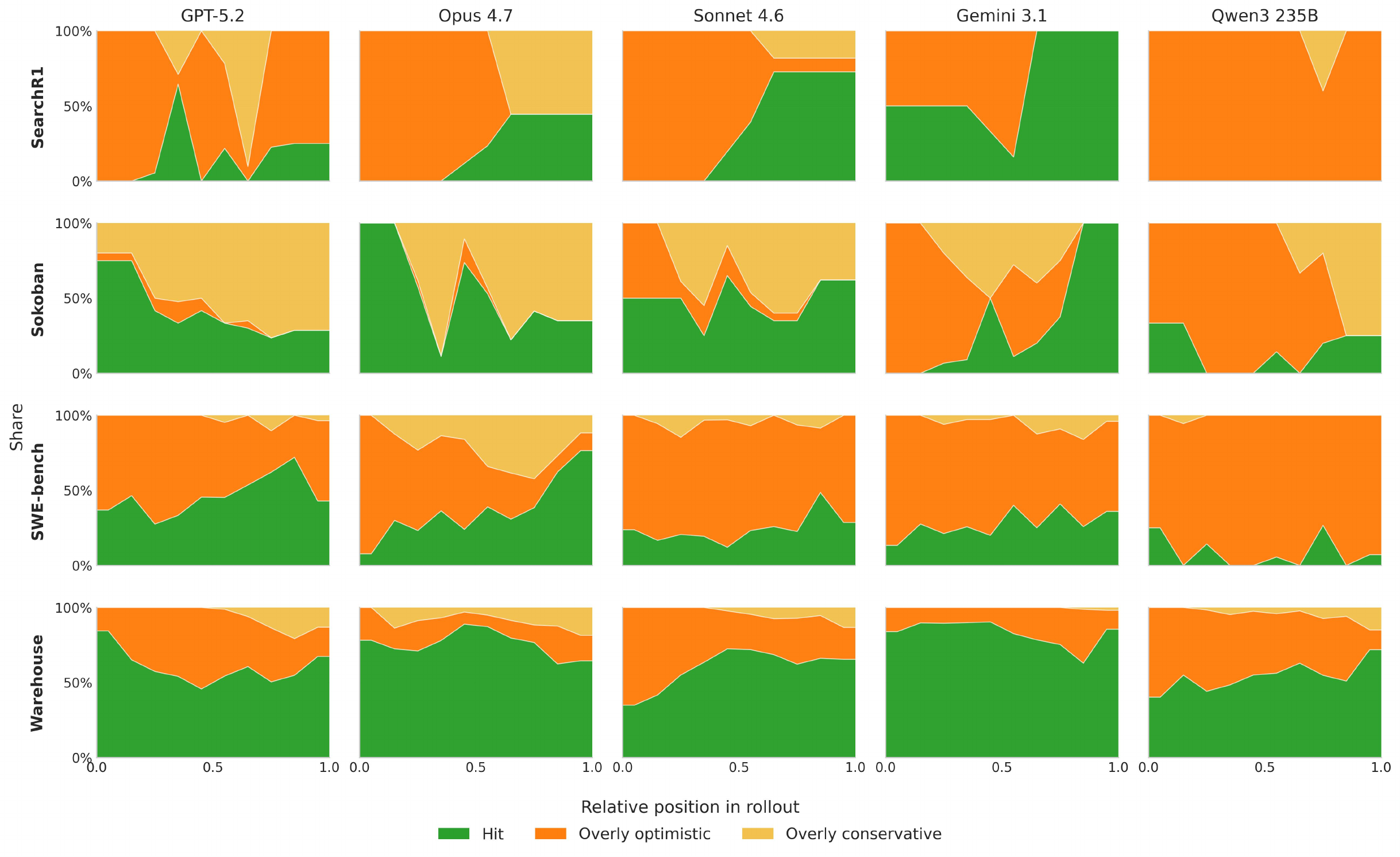}
    \caption{Optimistic misses dominate single-point budget estimates. Orange regions (predicted budget below realized) exceed yellow (predicted budget above realized) for most model, across all rollout-progress bins.}
    \label{fig:v2:optimism}
\end{figure}

Figure~\ref{fig:v2:optimism} analyzes the direction of estimation errors across rollout progress.
Across models, optimistic misses dominate conservative misses, meaning that agents more often underestimate the remaining budget than overestimate it.
This bias persists throughout execution, although conservative misses become more visible in later rollout stages.
Together, these results show that budget-estimation errors are not symmetric: models are systematically biased toward thinking the remaining task will be cheaper than it actually is.

Table~\ref{tab:v2:early-stop-perbench} reports the per-benchmark early-stopping tradeoff when trajectories are stopped at the first \texttt{impossible} prediction.
The results show that impossible predictions can save substantial budget on failed rollouts, especially in Warehouse and SWE-bench, while false-abort rates remain low for most model-benchmark pairs.
Search-R1 shows smaller savings because its rollouts are shorter, leaving less room for early stopping before the trajectory ends.
This supports the main result that budget estimates are actionable, but their utility depends on how early failure can be detected within each environment.

\begin{figure}[t]
    \centering
    \includegraphics[width=1\linewidth]{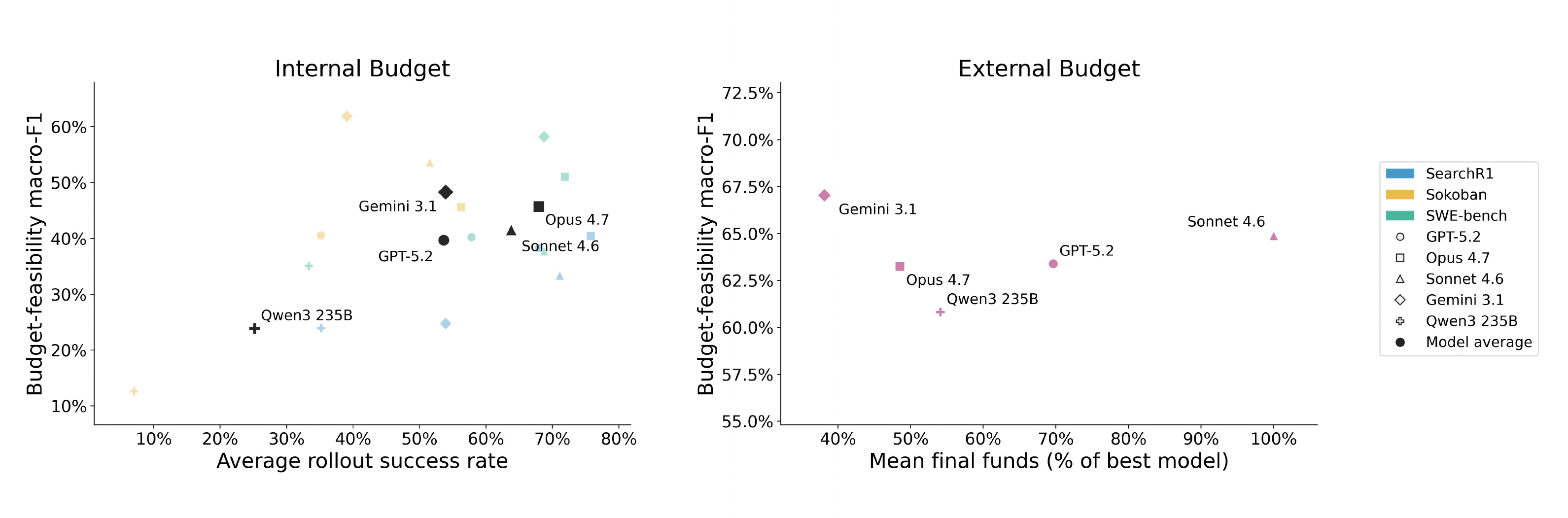}
    \caption{Progressive interval estimation evaluates an agent's budget-estimation ability separately from its task performance.
    The weak relationship between task success and estimation quality shows that doing the task well does not mean knowing how much budget remains.
    }
    \label{fig:v2:rollout-vs-feasibility}
    \vspace{-8pt}
\end{figure}

Figure~\ref{fig:v2:rollout-vs-feasibility} further separates task success from budget-estimation quality.
Across benchmarks and models, higher rollout success does not necessarily imply better feasibility prediction.
This shows that budget awareness is not simply a byproduct of stronger task-solving ability, but a distinct capability that must be evaluated directly.

\begin{table}[htbp]
\centering
\scriptsize
\caption{Per-benchmark early-stopping tradeoff by estimator model. False-abort prediction rate, failed-rollout token share saved, false-abort counts, and the number of failed rollouts stopped at least once.}
\label{tab:v2:early-stop-perbench}
\begin{tabular}{llrrrr}
\toprule
Benchmark & Model & False-abort rate & Saved tokens & False-abort count & Stopped failed rollouts \\
\midrule
SearchR1 & GPT-5.2 & 0.0\% & 0.0\% & 0 / 324 & 0 / 41 \\
SearchR1 & Opus 4.7 & 0.0\% & 3.5\% & 0 / 87 & 1 / 31 \\
SearchR1 & Sonnet 4.6 & 1.0\% & 0.0\% & 1 / 102 & 0 / 37 \\
SearchR1 & Gemini 3.1 & 1.2\% & 0.0\% & 1 / 85 & 0 / 59 \\
SearchR1 & Qwen3 235B & 0.0\% & 19.4\% & 0 / 1{,}079 & 40 / 83 \\
\addlinespace[2pt]
Sokoban & GPT-5.2 & 1.1\% & 30.5\% & 7 / 661 & 48 / 83 \\
Sokoban & Opus 4.7 & 0.0\% & 10.8\% & 0 / 355 & 9 / 56 \\
Sokoban & Sonnet 4.6 & 1.5\% & 27.9\% & 6 / 400 & 29 / 62 \\
Sokoban & Gemini 3.1 & 0.3\% & 48.0\% & 1 / 389 & 53 / 78 \\
Sokoban & Qwen3 235B & 0.1\% & 12.8\% & 1 / 1{,}038 & 33 / 119 \\
\addlinespace[2pt]
SWE-bench & GPT-5.2 & 0.0\% & 19.5\% & 0 / 512 & 12 / 27 \\
SWE-bench & Opus 4.7 & 0.0\% & 29.1\% & 0 / 512 & 7 / 18 \\
SWE-bench & Sonnet 4.6 & 0.0\% & 45.2\% & 0 / 512 & 8 / 20 \\
SWE-bench & Gemini 3.1 & 0.0\% & 50.7\% & 0 / 512 & 12 / 20 \\
SWE-bench & Qwen3 235B & 0.0\% & 7.0\% & 0 / 512 & 5 / 40 \\
\addlinespace[2pt]
Warehouse & GPT-5.2 & 13.8\% & 90.9\% & 176 / 1{,}279 & 64 / 64 \\
Warehouse & Opus 4.7 & 3.9\% & 41.9\% & 50 / 1{,}280 & 45 / 64 \\
Warehouse & Sonnet 4.6 & 5.4\% & 74.3\% & 69 / 1{,}280 & 64 / 64 \\
Warehouse & Gemini 3.1 & 4.8\% & 70.5\% & 61 / 1{,}280 & 58 / 64 \\
Warehouse & Qwen3 235B & 14.8\% & 85.4\% & 189 / 1{,}280 & 62 / 64 \\
\bottomrule
\end{tabular}
\end{table}

\section{Detailed Settings for Estimation RL and SFT}

\label{app:sft-rl-settings}
The SFT and RL experiments use Qwen/Qwen2.5-7B-Instruct as the base model.
The probe task is Sokoban 6x6 with one box and a 2500-token budget.
The SFT training set contains 794 samples.
The balanced test set contains 380 samples, split evenly between possible and impossible states.

The raw rollouts are first converted into alternating user/assistant message histories.
For each trajectory, a probe is inserted before the first action and after every completed turn.
At probe-after-turn $t$, the prompt contains the Sokoban system prompt and the conversation prefix up to $t$.
For $t=0$, the prompt contains the system prompt and the initial grid only.
The completed-turn token usage is computed from the visible user messages and the recorded API output-token counts, so that post-processed assistant text does not undercount hidden generation cost.
The true remaining-token label $r$ is the sum of all future assistant output tokens and all future user input tokens after the visible prefix.
A probe is labeled possible only if the original rollout succeeds and the visible-prefix tokens plus $r$ fit within the budget; otherwise it is labeled \texttt{impossible}.
The ablation data use a deterministic trajectory split with seed 42: 40\% for SFT, 50\% for RL, and 10\% for held-out evaluation.
Training splits are balanced with a 1:1 possible/impossible ratio after duplicating possible samples twice; trivial samples with $r=0$ are removed.

\textbf{SFT targets.}
SFT turns each probe into a full chat record by appending the ground-truth assistant answer.
The main runs use the no-thinking output template, so the assistant target is either \texttt{<answer>impossible</answer>} or \texttt{<answer>[L, H]</answer>}.
For a possible probe with remaining-token label $r$, the percentage-width targets are
\[
L=\max(1,\lfloor r(1-w)\rfloor), \qquad H=\lfloor r(1+w)\rfloor,
\]
with $w\in\{0.1,0.3,0.5\}$.
The fixed-width targets are
\[
L=\max(1,\lfloor r-w\rfloor), \qquad H=\lfloor r+w\rfloor,
\]
with $w\in\{100,500,1000\}$ tokens.
We also prepared a point-estimation SFT variant that predicts one integer, but the reported RL interval experiments warm start from interval SFT checkpoints.
All SFT runs use FSDP training for 5 epochs with learning rate $5\times 10^{-6}$, global batch size 16, and per-GPU micro-batch size 2.
The default SFT script sets maximum sequence length to 9216, while the v3 batch launcher overrides it to 16384.
Checkpoints are converted after epochs 2, 3, and 5.

\textbf{Training-name convention.}
Run names encode the supervision target and checkpoint used for evaluation.
\texttt{pct10}, \texttt{pct30}, and \texttt{pct50} denote interval SFT targets whose half-width is 10\%, 30\%, or 50\% of the true remaining-token label.
\texttt{fix100}, \texttt{fix500}, and \texttt{fix1000} denote interval SFT targets with fixed half-widths of 100, 500, or 1000 tokens.
The suffixes \texttt{e2}, \texttt{e3}, and \texttt{e5} identify checkpoints after 2, 3, and 5 SFT epochs.
An entry such as \texttt{SFT pct30 e5 $\rightarrow$ RL KL 0.05} means that the epoch-5 \texttt{pct30} SFT checkpoint initializes GRPO, with actor KL coefficient 0.05.
\texttt{Zero RL} denotes GRPO started directly from the base model, without SFT warm start.
\texttt{no think} and \texttt{+ think} distinguish whether the output template omits or includes a \texttt{<think>...</think>} field.
\texttt{scalar bug} denotes an invalid zero-RL run that learned scalar outputs instead of the required interval format.

\textbf{RL setup.}
RL is applied with GRPO on top of selected SFT checkpoints.
The RL parquet records contain the same prompt without the assistant answer, together with the rule-reward ground truth and the numeric \texttt{remaining\_tokens} field.
The final GRPO runs use 8 GPUs, global batch size 64, 16 rollouts per prompt, learning rate $5\times 10^{-7}$, 5 epochs, maximum prompt length 8192, maximum response length 1024, entropy coefficient 0, and no KL term inside the reward.
The actor still uses a low-variance KL loss against the reference policy; the reported sweep uses KL coefficients 0.05 and 0.1.
Evaluation decodes with vLLM at temperature 0 and a 512-token response cap.

\textbf{RL reward.}
The rule reward first extracts the content inside \texttt{<answer>...</answer>}.
Outputs without this tag, malformed intervals, reversed intervals, and scalar predictions receive zero reward for possible probes.
Impossible probes receive reward 0.2 only when the extracted answer is exactly \texttt{impossible}.
For a possible probe with true remaining tokens $r$, an interval $[L,H]$ receives reward only when $L \le r \le H$:
\[
R = 1.8 \cdot \max\left(0, 1 - \frac{H-L}{r}\right).
\]
This reward favors tight intervals that still cover the true value.
The 1.8-to-0.2 weighting discourages the all-impossible policy while keeping impossible-state recognition in the objective.

\section{Additional Data for RL}

Table~\ref{tab:appendix-sft-rl-results} reports the full SFT/RL sweep.
The zero-RL runs show that RL from scratch does not reliably learn the interval format.
Both no-thinking and thinking variants collapse to all-impossible answers.
The scalar-bug run has high classification accuracy, but it does not produce valid intervals.
These results show that SFT is needed before RL.

\begin{table}[htbp]
\centering
\scriptsize
\setlength{\tabcolsep}{3.5pt}
\caption{Sokoban budget-probe results for the SFT and RL sweep. Acc. is feasible/impossible classification accuracy; Rec. pos/imp are recalls for feasible and impossible probes; Cover is interval coverage on feasible probes; MRE is midpoint relative error; R is the combined reward.}
\label{tab:appendix-sft-rl-results}
\resizebox{\textwidth}{!}{%
\begin{tabular}{lrrrrrrr}
\toprule
Model & Acc. & Rec. pos & Rec. imp & Cover & MRE P50 & MRE P90 & R \\
\midrule
Qwen2.5-7B-Instruct & 25.5\% & 46.3\% & 4.7\% & 10.5\% & 65.5\% & 277.2\% & 0.021 \\
Zero RL scalar bug & 90.5\% & 93.2\% & 87.9\% & 0.0\% & 0.0\% & 0.0\% & 0.088 \\
Zero RL no think & 50.0\% & 0.0\% & 100.0\% & 0.0\% & 0.0\% & 0.0\% & 0.100 \\
Zero RL + think & 50.0\% & 0.0\% & 100.0\% & 0.0\% & 0.0\% & 0.0\% & 0.100 \\
SFT pct10 e2 & 87.6\% & 100.0\% & 75.3\% & 26.3\% & 45.1\% & 84.0\% & 0.265 \\
SFT pct10 e3 & 89.5\% & 95.3\% & 83.7\% & 19.5\% & 62.0\% & 135.4\% & 0.224 \\
SFT pct10 e5 & 90.3\% & 93.7\% & 86.8\% & 26.3\% & 45.1\% & 84.2\% & 0.277 \\
SFT pct30 e2 & 88.2\% & 99.5\% & 76.8\% & 39.5\% & 53.5\% & 86.7\% & 0.204 \\
SFT pct30 e3 & 90.0\% & 96.3\% & 83.7\% & 38.4\% & 51.8\% & 84.3\% & 0.205 \\
SFT pct30 e5 & 91.1\% & 95.3\% & 86.8\% & 37.4\% & 53.9\% & 86.4\% & 0.210 \\
SFT pct50 e2 & 86.6\% & 100.0\% & 73.2\% & 41.1\% & 57.9\% & 848.1\% & 0.094 \\
SFT pct50 e3 & 89.7\% & 96.3\% & 83.2\% & 54.2\% & 48.0\% & 86.0\% & 0.099 \\
SFT pct50 e5 & 90.5\% & 94.7\% & 86.3\% & 35.8\% & 55.2\% & 303.4\% & 0.101 \\
SFT fix100 e2 & 86.6\% & 100.0\% & 73.2\% & 53.7\% & 48.6\% & 83.7\% & 0.085 \\
SFT fix100 e3 & 90.5\% & 97.4\% & 83.7\% & 52.6\% & 42.9\% & 83.6\% & 0.095 \\
SFT fix100 e5 & 90.0\% & 92.6\% & 87.4\% & 46.8\% & 49.1\% & 88.4\% & 0.105 \\
SFT fix500 e2 & 86.8\% & 99.5\% & 74.2\% & 90.0\% & 65.9\% & 427.3\% & 0.074 \\
SFT fix500 e3 & 90.0\% & 96.8\% & 83.2\% & 85.3\% & 68.4\% & 294.0\% & 0.083 \\
SFT fix500 e5 & 90.3\% & 93.7\% & 86.8\% & 84.7\% & 67.9\% & 441.8\% & 0.087 \\
SFT fix1000 e2 & 86.3\% & 100.0\% & 72.6\% & 96.8\% & 182.7\% & 620.0\% & 0.073 \\
SFT fix1000 e3 & 90.0\% & 96.3\% & 83.7\% & 92.6\% & 169.6\% & 627.3\% & 0.084 \\
SFT fix1000 e5 & 90.0\% & 93.2\% & 86.8\% & 89.5\% & 169.5\% & 639.7\% & 0.087 \\
SFT pct10 e5 $\rightarrow$ RL KL 0.05 & 88.9\% & 90.0\% & 87.9\% & 26.8\% & 38.9\% & 84.0\% & 0.281 \\
SFT pct10 e5 $\rightarrow$ RL KL 0.1 & 89.2\% & 90.5\% & 87.9\% & 27.9\% & 38.9\% & 84.0\% & 0.289 \\
SFT fix100 e3 $\rightarrow$ RL KL 0.05 & 90.3\% & 93.7\% & 86.8\% & 13.2\% & 177.6\% & 350.0\% & 0.126 \\
SFT pct30 e5 $\rightarrow$ RL KL 0.05 & 90.0\% & 92.6\% & 87.4\% & 46.8\% & 28.2\% & 87.8\% & 0.264 \\
\bottomrule
\end{tabular}%
}
\vspace{0.5em}
\begin{minipage}{0.98\textwidth}
\footnotesize
\textit{Notes.}
\texttt{SFT} denotes supervised fine-tuning; \texttt{RL} denotes the subsequent GRPO stage; and \texttt{Zero RL} denotes GRPO from the base model without an SFT warm start.
\texttt{pct10}, \texttt{pct30}, and \texttt{pct50} are SFT target intervals with widths equal to 10\%, 30\%, or 50\% of the true remaining token count.
\texttt{fix100}, \texttt{fix500}, and \texttt{fix1000} are fixed-width target intervals of 100, 500, or 1000 tokens.
The suffixes \texttt{e2}, \texttt{e3}, and \texttt{e5} mark checkpoints after 2, 3, and 5 SFT epochs; \texttt{KL} is the RL KL-penalty coefficient.
\texttt{+ think} includes a short reasoning field before the answer, while \texttt{no think} omits it.
\texttt{scalar bug} is an invalid zero-RL run that outputs scalar values instead of required intervals.
\texttt{Acc.} is binary feasibility accuracy.
\texttt{Rec. pos} and \texttt{Rec. imp} are recall on possible and impossible states.
\texttt{Cover} is computed on truly possible states where the model outputs an interval $[L,H]$ and measures whether the true remaining count $r$ satisfies $L \leq r \leq H$.
\texttt{MRE} is midpoint relative error, $\mathrm{MRE}=|((L+H)/2)-r|/r$; \texttt{P50} and \texttt{P90} are its median and 90th percentile.
\texttt{R} is the average sparse reward used for RL evaluation.
\end{minipage}
\end{table}

\section{Prompts for Agent Estimation and SFT/RL}

This section summarizes the prompt format used for agent budget estimation.
All variants follow the same structure.
The prompt defines the budget unit, lists the context fields, and ends with strict output rules.
The compact templates below show the key instructions.

\textbf{SearchR1 token-budget prompt.}
\begin{center}
{\setlength{\fboxsep}{8pt}
\fcolorbox{orange!70!black}{orange!3}{%
\begin{minipage}{0.92\linewidth}
\small
\textbf{Based on the provided SearchQA rollout context, you are given the following:}
\begin{enumerate}
\item You have completed \texttt{\{completed\_turns\}} turns.
\item Each turn, your token consumption is \texttt{\{turn\_token\_usage\_text\}}.
\item You need to finish the task within \texttt{\{max\_context\_window\_tokens\}} total tokens.
\end{enumerate}

\textbf{Now, estimate:}
\begin{enumerate}
\item Whether the search agent can still finish the task successfully within \texttt{\{max\_context\_window\_tokens\}} total tokens (input + output).
\item If yes, how many additional tokens (input + output) are still needed to finish the task, starting from the next turn. Return an estimation interval: at least est\_low tokens and at most est\_high tokens.
\item If no, answer "impossible".
\item Prioritize the can-finish judgment over interval tightness. If you think the task can finish within budget, make the interval as tight as possible while still covering the true remaining token budget.
\end{enumerate}

\textbf{Example:}
For a three-turn interaction, suppose only Turn 1 has been completed.
The full interaction is:
Turn 1: input X1 tokens, output Y1 tokens;
Turn 2: input X2 tokens, output Y2 tokens;
Turn 3: input X3 tokens, output Y3 tokens.
You will receive:
turn\_token\_usage\_text: Turn 1: input X1 tokens, output Y1 tokens
You should estimate:
X2 + Y2 + X3 + Y3

\textbf{Output exactly one of the following:}

$<$\texttt{think}$>$\texttt{[YOUR THINKING]}$<$\texttt{/think}$>$$<$\texttt{answer}$>$\texttt{[est\_low, est\_high]}$<$\texttt{/answer}$>$

or

$<$\texttt{think}$>$\texttt{[YOUR THINKING]}$<$\texttt{/think}$>$$<$\texttt{answer}$>$\texttt{impossible}$<$\texttt{/answer}$>$
\end{minipage}}}
\end{center}

\textbf{Sokoban token-budget prompt.}
\begin{center}
{\setlength{\fboxsep}{8pt}
\fcolorbox{orange!70!black}{orange!3}{%
\begin{minipage}{0.92\linewidth}
\small
\textbf{Based on the provided rollout context, you are provided below information:}
\begin{enumerate}
\item You have completed \texttt{\{completed\_turns\}} turns.
\item Each turn, your token consumption is \texttt{\{turn\_token\_usage\_text\}}.
\item You need to finish the task within \texttt{\{max\_context\_window\_tokens\}} tokens.
\end{enumerate}

\textbf{Now, estimate:}
\begin{enumerate}
\item Whether you can finish the task successfully within \texttt{\{max\_context\_window\_tokens\}} total tokens (input + output).
\item If yes, how many additional tokens (input + output) are still needed to finish the task, starting from the next turn. Return an estimation interval: at least est\_low tokens and at most est\_high tokens.
\item If no, answer "impossible".
\item You should try your best to estimate whether the task can finish within budget (most important). If you think the task can finish within budget, your interval should be as tight as possible while still covering the true remaining token budget.
\end{enumerate}

\textbf{Example:}
For a three-turn interaction, suppose only Turn 1 has been completed.
The full interaction is:
Turn 1: input X1 tokens, output Y1 tokens;
Turn 2: input X2 tokens, output Y2 tokens;
Turn 3: input X3 tokens, output Y3 tokens.
You will receive:
turn\_token\_usage\_text: Turn 1: input X1 tokens, output Y1 tokens
You should estimate:
X2 + Y2 + X3 + Y3

\textbf{Output exactly one of the following:}

$<$\texttt{think}$>$\texttt{[YOUR THINKING]}$<$\texttt{/think}$>$$<$\texttt{answer}$>$\texttt{[est\_low, est\_high]}$<$\texttt{/answer}$>$

or

$<$\texttt{think}$>$\texttt{[YOUR THINKING]}$<$\texttt{/think}$>$$<$\texttt{answer}$>$\texttt{impossible}$<$\texttt{/answer}$>$
\end{minipage}}}
\end{center}

\textbf{SWE-bench token-budget prompt.}
\begin{center}
{\setlength{\fboxsep}{8pt}
\fcolorbox{orange!70!black}{orange!3}{%
\begin{minipage}{0.92\linewidth}
\small
\textbf{Based on the provided SWE-bench rollout context, you are given the following:}
\begin{enumerate}
\item The coding agent has completed \texttt{\{completed\_turns\}} turns.
\item Per-turn token usage so far, excluding reused history from earlier turns, is: \texttt{\{turn\_token\_usage\_text\}}.
\item The full task must finish within \texttt{\{max\_context\_window\_tokens\}} total tokens.
\end{enumerate}

\textbf{Estimate:}
\begin{enumerate}
\item Whether the agent can still finish the software issue successfully within \texttt{\{max\_context\_window\_tokens\}} total tokens.
\item If yes, how many additional tokens (input + output) are still needed from the next turn onward. Return an interval: at least est\_low tokens and at most est\_high tokens.
\item If no, answer "impossible".
\item Prioritize the can-finish judgment over interval tightness. If the task still looks finishable, keep the interval as tight as possible while still covering the true remaining token budget.
\end{enumerate}

Think about typical SWE-bench costs such as repository inspection, targeted code edits, running validation commands, reading failures, and one or two repair iterations.

\textbf{Output exactly one of the following:}

$<$\texttt{think}$>$ \texttt{[YOUR THINKING]} $<$\texttt{/think}$>$$<$\texttt{answer}$>$\texttt{[est\_low, est\_high]}$<$\texttt{/answer}$>$

or

$<$\texttt{think}$>$ \texttt{[YOUR THINKING]} $<$\texttt{/think}$>$$<$\texttt{answer}$>$\texttt{impossible}$<$\texttt{/answer}$>$
\end{minipage}}}
\end{center}

\textbf{External Warehouse prompt.}
\begin{center}
{\setlength{\fboxsep}{8pt}
\fcolorbox{orange!70!black}{orange!3}{%
\begin{minipage}{0.92\linewidth}
\small
\textbf{System prompt.}
You are an evaluation agent for historical warehouse-management rollouts.
Determine whether the rollout can still finish successfully within the remaining resource budgets while reaching the target cash threshold.
If it can, estimate the remaining time, warehouse cumulative occupancy, and cumulative cost still needed from the next turn onward.
Follow the required output format exactly.

\medskip
\textbf{Based on the provided warehouse rollout context, you are given the following information:}
\begin{enumerate}
\item You have completed \texttt{\{completed\_weeks\}} weeks in \texttt{\{completed\_turns\}} turns.
\item Current cumulative usage so far:
\begin{itemize}
\item time\_weeks: \texttt{\{current\_time\_weeks\}}
\item warehouse\_item\_weeks: \texttt{\{current\_warehouse\_item\_weeks\}}
\item cumulative\_cost\_usd: \texttt{\{current\_cost\_usd\}}
\end{itemize}
\item Current cash is \texttt{\{current\_cash\_usd\}} USD. To count as finished, final cash must reach at least \texttt{\{target\_cash\_usd\}} USD.
\item Historical resource consumption by completed step is:
\texttt{\{resource\_consumption\_text\}}
\item The rollout must finish within all three budgets:
\begin{itemize}
\item time\_weeks $<$= \texttt{\{budget\_time\_weeks\}}
\item warehouse\_item\_weeks $<$= \texttt{\{budget\_warehouse\_item\_weeks\}}
\item cumulative\_cost\_usd $<$= \texttt{\{budget\_cost\_usd\}}
\end{itemize}
\end{enumerate}

\textbf{Now, estimate:}
\begin{enumerate}
\item Whether the rollout can still finish successfully within all three budgets while also reaching the target cash.
\item If yes, how much additional usage is still needed from the next turn onward. Return one interval for each metric.
\item If no, answer "impossible".
\item Prioritize the can-finish judgment over interval tightness. If you think the rollout can finish within budget, make each interval as tight as possible while still covering the true remaining value.
\end{enumerate}

\textbf{Output exactly one of the following:}

$<$\texttt{think}$>$\texttt{[YOUR THINKING]}$<$\texttt{/think}$>$$<$\texttt{answer}$>$\texttt{time\_weeks:[est\_low, est\_high], warehouse\_item\_weeks:[est\_low, est\_high], cumulative\_cost\_usd:[est\_low, est\_high]}$<$\texttt{/answer}$>$

or

$<$\texttt{think}$>$\texttt{[YOUR THINKING]}$<$\texttt{/think}$>$$<$\texttt{answer}$>$\texttt{impossible}$<$\texttt{/answer}$>$
\end{minipage}}}
\end{center}

\textbf{Sokoban SFT/RL budget-probe prompt.}
\begin{center}
{\setlength{\fboxsep}{8pt}
\fcolorbox{orange!70!black}{orange!3}{%
\begin{minipage}{0.92\linewidth}
\small
\textbf{System prompt.}
You're a helpful assistant. You are solving the Sokoban puzzle.
Push all boxes to targets.
You are given the grid and zero-indexed coordinates of the player, boxes, and targets.
You can push but not pull boxes, and cannot push a box through a wall.
Your available actions are:
Up, Down, Left, Right.
You may output at most 3 action(s) in a single turn, separated by the action separator " || ".

\medskip
\textbf{User estimation prompt.}

\textbf{Based on the provided rollout context, you are provided below information:}
\begin{enumerate}
\item You have completed \texttt{\{completed\_turns\}} turns.
\item Each turn, your token consumption is \texttt{\{turn\_token\_usage\_text\}}.
\item You need to finish the task within \texttt{\{max\_context\_window\_tokens\}} tokens.
\end{enumerate}

\textbf{Now, estimate:}
\begin{enumerate}
\item Whether you can finish the task successfully within \texttt{\{max\_context\_window\_tokens\}} total tokens (input + output).
\item If yes, how many additional tokens (input + output) are still needed to finish the task, starting from the next turn. Return an estimation interval: at least est\_low tokens and at most est\_high tokens.
\item If no, answer "impossible".
\item You should try your best to estimate whether the task can finish within budget (most important). If you think the task can finish within budget, your interval should be as tight as possible while still covering the true remaining token budget.
\end{enumerate}

\textbf{Example:}
For a three-turn interaction, suppose only Turn 1 has been completed.
The full interaction is:
Turn 1: input X1 tokens, output Y1 tokens;
Turn 2: input X2 tokens, output Y2 tokens;
Turn 3: input X3 tokens, output Y3 tokens.
You will receive:
turn\_token\_usage\_text: Turn 1: input X1 tokens, output Y1 tokens
You should estimate:
X2 + Y2 + X3 + Y3

\textbf{Output exactly one of the following:}

$<$\texttt{answer}$>$\texttt{[est\_low, est\_high]}$<$\texttt{/answer}$>$

or

$<$\texttt{answer}$>$\texttt{impossible}$<$\texttt{/answer}$>$
\end{minipage}}}
\end{center}



\end{document}